%% file: main.tex
\documentclass[10pt]{article} 

\usepackage{algorithm}
\usepackage{algpseudocode}
\usepackage{graphicx}
\usepackage{subfig}
\usepackage{makecell}
\usepackage{booktabs} 
\usepackage{multirow}
\usepackage{adjustbox}

\usepackage{xcolor}
\usepackage{enumitem}
\setlist{nosep, leftmargin=1em}

\usepackage[preprint]{tmlr}

\input{math_commands.tex}

\usepackage{hyperref}
\usepackage{url}

\title{Learning Using a Single Forward Pass}

\author{\name Aditya Somasundaram \thanks{AS \& PM: joint first authors (order by coin toss); AB: PI. Work partly done at IIT Hyderabad.} \email as7458@columbia.edu \\
    \addr Columbia University
      \AND
      \name Pushkal Mishra\footnotemark[1] \email pumishra@ucsd.edu\\
      \addr University of California San Diego
      \AND
      \name Ayon Borthakur\footnotemark[1]  \email ayon.borthakur@iitg.ac.in\\
      \addr IIT Guwahati}

\def\month{06}  
\def\year{2025} 
\def\openreview{\url{https://openreview.net/forum?id=EDQ8QOGqjr}} 

\begin{document}

\maketitle


\begin{abstract}
We propose a learning algorithm to overcome the limitations of traditional backpropagation in resource-constrained environments: Solo Pass Embedded Learning Algorithm (SPELA). SPELA operates with local loss functions to update weights, significantly saving on resources allocated to the propagation of gradients and storing computational graphs while being sufficiently accurate. Consequently, SPELA can closely match backpropagation using less memory. Moreover, SPELA can effectively fine-tune pre-trained image recognition models for new tasks. Further, SPELA is extended with significant modifications to train CNN networks, which we evaluate on CIFAR-10, CIFAR-100, and SVHN 10 datasets, showing equivalent performance compared to backpropagation. Our results indicate that SPELA, with its features such as local learning and early exit, is a potential candidate for learning in resource-constrained edge AI applications.
\end{abstract}

\section{Introduction}
\label{sec:intro}

Backpropagation (BP) is a long-standing, fundamental algorithm for training deep neural networks (NNs) \citep{werbos, Rumelhart1986, lecun1989, LeCun2015}. It is widely used in training multi-layer perceptrons (MLP), convolutional neural networks (CNN), recurrent neural networks (RNN), and now transformers. It is a loss minimization problem involving thousands, if not millions, of parameters. Data is first propagated through the network using forward passes, and the entire computational graph is stored. The difference (error) between the final layer output and the label is utilized to update the weight matrices of the network. The error is then propagated backward from the final layer using the chain rule of differentiation, which uses the stored computational graph to compute the associated gradients for each layer. 


Careful observations of backpropagation drive home the point of no free lunch. Backpropagation works within a global learning framework, i.e., a single weight update requires knowledge of the gradient of every parameter (global/non-local learning problem) \citep{WHITTINGTON2019235}. Backpropagation requires the storage of neural activations computed in the forward pass for use in the subsequent backward pass (weight transport problem) \citep{ lillicrap2014random, akrout}. For every forward pass, the backward pass is computed with the forward-pass updates frozen, which prevents online utilization of inputs (update locking problem) \citep{czarnecki, jaderberg17a}. Furthermore, there are several differences between backpropagation and biological learning, as mentioned in Section \ref{sec:extended-intro}.



The constraints of backpropagation lead to significant challenges, such as high memory requirements. These limitations render backpropagation computationally expensive and unsuitable for applications in resource-constrained scenarios \citep[example: on-device machine learning (ODL)][]{tinyTL, zhu2023ondevice}. Efforts to mitigate these algorithmic constraints of backpropagation can lead to improved learning efficiency.

We introduce and investigate a multi-layer neural network training algorithm — SPELA (Solo Pass Embedded Learning Algorithm) that uses embedded vectors as priors to preserve data structure as it passes through the network. 
Previous studies indicate that prior knowledge helps one learn faster and more easily \citep{goyal, wang2023theoretical}. Although we do not claim complete biological plausibility, SPELA is built on the premise that biological neural networks utilize local learning \citep{illing2021local} and neural priors to form representations. We introduce neural priors as symmetric vectors distributed on a high-dimensional sphere whose dimension equals the size of the corresponding neuron layer. Our back-propagation-free learning algorithm demonstrates a significant gain in computational efficiency, making it suitable for on-device learning (ODL) and brain-inspired learning features, such as early exit (for layered cognitive reasoning \citep{Scardapane_2020}) and local learning. In this paper, we make the following contributions:

\begin{itemize}

    \item \textbf{Design:} We introduce SPELA. Next, we extend SPELA to convolutional neural networks. SPELA is a family of algorithms that use a single forward pass (with no backward pass) for training. During inference, output from any layer can be utilized for prediction. It makes an innovative use of embedded vectors as neural priors for efficient learning.

    \item \textbf{Evaluate:} Experiments conducted in this paper indicate that SPELA closely matches backpropagation in performance, and due to its computational efficiency, maintains an edge over it in resource-constrained scenarios. Moreover, SPELA can efficiently fine-tune models trained with backpropagation (transfer learning). In addition, extending SPELA to convolutional neural networks (CNNs) allows for complex image classification.

    \item \textbf{Complexity Analysis:} Theoretical bounds for peak memory usage show that SPELA can edge over backpropagation in the analyzed settings. 
\end{itemize}

\section{Related works}
\label{sec:extended-intro}

Recently, there has been significant interest in designing efficient training methods for multi-layer neural networks. \citet{hinton2022forwardforward} presents the Forward-Forward (FF) algorithm for neural network learning. In FF, backpropagation is replaced with two forward passes: one with positive (real) data and the other using generated negative data. Each layer aims to optimize a goodness metric for positive data and minimize it for negative data. Separating positive and negative passes in time enables offline processing, facilitating image pipelining without activity storage or gradient propagation interruptions. These algorithms have garnered significant attention, and multiple modifications have been proposed in conjunction with applications in image recognition  \citep{lee2023symba, Pau, AliMomeni2023, dooms2024the, chen2024selfcontrastiveforwardforwardalgorithm} and in graph neural networks \citep{park2024forward}. In our experiments, SPELA can classify any layer without storing goodness memory, in contrast to the FF approach, which aggregates goodness values across layers. Furthermore, unlike FF, SPELA eliminates the need to generate separate negative data for training. Instead, it efficiently uses the available data without requiring additional processing. 

Using forward and backward passes, \citet{pehlevan} introduced the concept of a non-negative similarity matching cost function for spiking neural networks to exhibit local learning and enable effective use of neuromorphic hardware. \citet{Lansdell2020Learning} introduced a hybrid learning approach wherein each neuron learns to approximate the gradients. The feedback weights provide a biologically plausible way to achieve performance comparable to networks trained via backpropagation. \citet{giampaolo} follows a similar strategy to our approach of dividing the entire network into sub-networks and training them locally using backpropagation (SPELA divides the network into sequential layers). Rather than backward propagation, \citet{dellaferrera22a} uses two forward passes and uses the global error to modulate the second forward-pass input. Recently, \citet{li2025noproptrainingneuralnetworks} proposed a diffusion-inspired denoising-based training of neural networks. We consider \citet{dellaferrera22a} as the closest match to our approach, but with one forward pass and layer-specific non-global error during training for SPELA. Inspired by pyramidal neurons, \citet{lv2025dendriticlocalizedlearningbiologically} used distinct weights for forward and backward passes to train a multi-layer network in a biologically plausible manner. As detailed by \citet{Pau, srinivasan2024forward}, these algorithms still lack several desired characteristics of on-device learning. 



\section{Methods}
\label{sec:methods}

\subsection{Network Initialization and Learning Methods}
The network is defined as follows: there are $L$ layers, each containing $l_i$ neurons followed by a nonlinear activation function (e.g Leaky ReLU). The weights of the network are initialized randomly. Each layer $L_i$ (except the input layer) has $N$ (number of classes in the given dataset) number of symmetric vectors, each of dimension $l_i$. These symmetric vectors are assigned a unique class. As the activation vector is also in the $l_i$ dimensional space, we can measure how close the activation vector points to a particular symmetric vector using a simple cosine similarity function \citep{AliMomeni2023}. Based on cosine similarity, the network predicts the class assigned to the symmetric vector closest to the activation vector. These symmetric vectors remain fixed and are not updated during training. We describe the symmetric vector generation method in detail in Section \ref{sec: sym emb}. 

\begin{figure*}[!htb]
    \centering
    \subfloat[Network Architecture]{%
        \includegraphics[width=9.25cm]{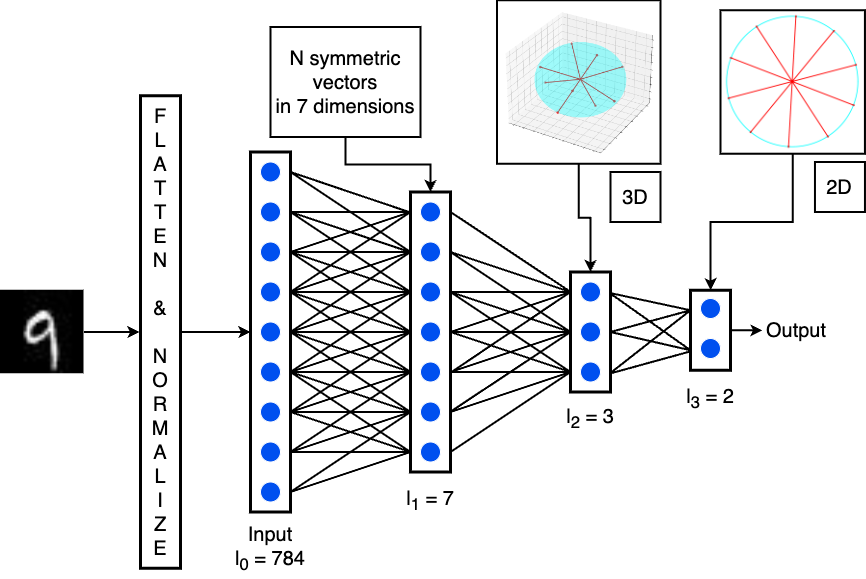}
        \label{fig:net_arch}%
        }%
    \hspace{1.5mm}%
    \subfloat[Prediction Method]{%
        \includegraphics[width=3.25cm]{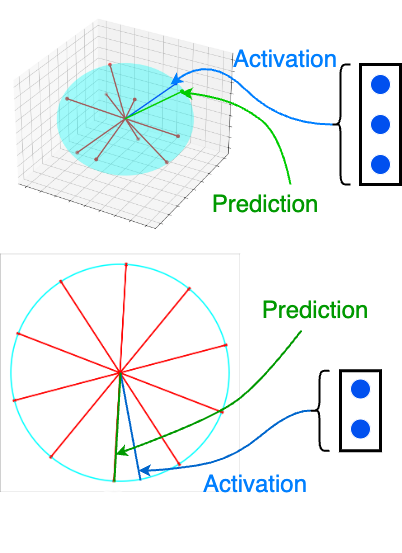}
        \label{fig:pred_method}
        }%
    \caption{(a) Network Architecture: Each layer possesses a distinct set of symmetric vectors. Here, the network is trained on MNIST-10, resulting in 10 symmetric vectors. (b) Prediction Method: Inference is performed using the closeness of activation and symmetric vectors. The activation is represented in blue, and the prediction is in green.}
    \label{fig:mlp_net_arch}
\end{figure*}

\subsection{SPELA description}
\label{sec: spela_description}

Algorithm \ref{alg:train_mlp_spela_new} and \ref{alg:alg_infer} describe SPELA's training and testing methodologies, respectively. In this method, after passing a data point through layer $i$, we update layer $i$’s weights and biases and then propagate the data point to layer $i+1$. Each layer updates its parameters as the data moves through the network using its local loss function. In this fashion, parallel training of all $n$ layers becomes possible. Table \ref{tab:passes} contrasts SPELA and other learning algorithms. Our algorithm exhibits the most favorable traits for the applications discussed in this work: a single forward pass for training, no backward pass, and a local loss function with no storage of activations.

\begin{table}[!htb]
    \centering
    \begin{tabular}{|c| c| c| c| c| c|}
        \hline
        \textbf{Learning Methods} & \textbf{BP} & \textbf{FF} & \textbf{PEP} &\textbf{MPE} & \textbf{SPELA} \\
        \hline
         Forward Pass & 1 & 2 & 2 & 3 & 1 \\
         \hline
         Backward Pass & 1 & 0 & 0 & 0 & 0 \\
         \hline
         Weight Update & 1 & 2 & 1 & 1 & 1 \\
         \hline
         Loss function & global & local & global & global & local \\
         \hline
         Activations & all & current & all & current & current \\
         \hline
    \end{tabular}
    \caption{Different learning algorithms are compared and contrasted with SPELA. PEP stands for PEPITA \citep{dellaferrera22a} and MPE for MEMPEPITA \citep{Pau}.}
    \label{tab:passes}
\end{table}

\begin{algorithm}[!htb]
\caption{\textbf{Training MLP with SPELA}}\label{alg:train_mlp_spela_new}
\begin{algorithmic}[1]
\State \textbf{Given:} An input (X), label (l), number of layers (K), and number of epochs (E)
\State \textbf{Define:} $\text{cos\textunderscore sim}(A, B) = \frac{A . B}{||A|| . ||B||}$ and $\text{normalize}(X) = \frac{X}{||X||}$ \Comment{Dot product and normalization of vector}
\State \textbf{Set:} $h_0 = x$
\For{$e \gets 1$ to $E$} \Comment{Iterate through epochs}
    \For{$k \gets 1$ to $K$} \Comment{Iterate through layers}
        \State $h_{k-1}$ = normalize($h_{k-1}$)
        \State $h_k = \sigma_k (W_k h_{k-1} + b_k)$
        \State $\text{loss}_k = - \text{cos\textunderscore sim}(h_k, \text{vecs}_k(l))$ \Comment{$\text{vecs}_k$(.) is the set of symmetric vectors}
        \State $W_k \leftarrow W_k - \alpha * \nabla_{W_k}(\text{loss}_k)$ \Comment{Weight update using local loss}
        \State $b_k \leftarrow b_k - \alpha * \nabla_{b_k}(\text{loss}_k)$ \Comment{Bias update using local loss}
    \EndFor
\EndFor
\end{algorithmic}
\end{algorithm}

\subsection{Complexity Analysis}
\label{sec: complexity analysis}

Updating the weights of the final layer in backpropagation requires one matrix-vector multiplication. Every other layer requires two matrix-vector multiplications to consider gradients from subsequent layers. SPELA can be visualized as cascading blocks of one-layer networks that perform classification at every junction. It can be viewed as a sequence of final layers from the backpropagation algorithm. Updating the weights of any layer using SPELA (as the weight updates are all analogous to the final layer of backpropagation) requires only one matrix-vector multiplication. For deep neural nets, the number of vector-matrix multiplications needed for weight updates is half of what backpropagation requires. Furthermore, this excludes other operations required by backpropagation and not by SPELA, such as transposing of weights. 
Both algorithms would need one vector-matrix multiplication for a forward pass. Considering this, SPELA requires about 0.67 as many matrix-vector multiplications as backpropagation (where the relative MACC for training is twice that of inference \citep{tinyTL}). Regarding memory (see Table \ref{tab:theory-res}), we describe the complexities involved in variables that must be saved to calculate the weight updates. We do not include overhead memory complexities from storing weights, optimizer states, temporary variables, or other sources. At each computational step, SPELA trains only a single layer; hence, only that layer's activation needs to be stored. SPELA offers superior computational and memory complexity compared to backpropagation. 



\begin{table}[!htb]
    \centering
    \begin{tabular}{|c| c| c| c|}
        \hline
        \textbf{Algorithm} & \makecell{\textbf{Forward pass}\\ \textbf{complexity}} & \makecell{\textbf{Weight update}\\ \textbf{complexity}} & \makecell{\textbf{Memory} \\ \textbf{complexity}} \\
         \hline
         SPELA & $N^2 L$ & $L N^2$ & $N$ \\
         \hline
         BP & $N^2 L$ & $2 L N^2$ & $L N$ \\
         \hline
    \end{tabular}
    \caption{The computation and memory complexities are shown above. The NN is considered to have L layers, each having N neurons. The complexity of a vector matrix multiplication is assumed to be $O(N^2)$. Here, for SPELA, it is assumed that activations are not stored.}
    \label{tab:theory-res}
\end{table}




\subsection{The learning in SPELA}
\label{sec: theo_spela}

During SPELA learning, every layer can be considered a classifier head. Instead of moving the activation towards a one-hot encoded vector as in conventional training schemes, we focus on moving the activation vector towards a predefined symmetric vector. At each layer, output $o_i = \frac{h_{i}}{||h_{i}||}$, where $h_i = \sigma(z_i)$, where $z_i = W_i h_{i-1} + b_i$, where $W_{i}$ and $b_{i}$ are layer weights and bias, $h_{i-1}$ is the previous layer activation output,  $\sigma()$ is an activation such as Leaky ReLU. Also, assume that the symmetric vector assigned to the correct class is $v$. Then the loss is defined as the negative of cosine similarity: $L = -o_i^T \cdot v$, as $||o_i|| = ||v|| = 1$. Mathematically, we compute 

\[
\frac{\partial L}{\partial o_i} = \frac{\partial }{\partial o_i}{-o_i^T \cdot v} = -v \equiv g_o
\text{ and }
\frac{\partial L}{\partial h_i} = \frac{\partial o_i}{\partial h_i} g_o = \frac{1}{||h_i||}(I - o_i o_i^T) \equiv g_h
\]

\[
\frac{\partial L}{\partial z_i} = g_h \odot \sigma'(z_i) \equiv g_z \text{, where } \sigma'(z) = \begin{cases}
    1\quad z > 0,\\
    \alpha\quad z \le 0
\end{cases}
\]
\[
\frac{\partial L}{\partial W_i} = g_z h_{i-1}^T \text{ and } \frac{\partial L}{\partial b_i} = g_z
\]

The same process is identically applicable to all layers $L$ in the network. This finally results in
$W_{i}^{t} \leftarrow W_{i}^{t-1} - \gamma g_z h_{i-1}^T \text{ and } b_{i}^{t} \leftarrow b_{i}^{t-1} - \gamma g_z, \text{where $\gamma$ is the learning rate and $t$ is iteration step.}$ In our implementation, the symmetric vector embeddings represent the layer-wise classifier head weights, which emulate cosine similarity computation, without getting updated during training. The output of this classifier head is utilized to compute a cross-entropy loss or a cosine loss ($\log$(2 - cosine similarity). The derivation above assumes a simpler setting, allowing readers to understand weight updates.

\subsection{SPELA for Convolutional Neural Network}
For the Convolutional Neural Networks (CNNs) (Figure \ref{fig:cnn_net_arch}), we use the interleaving of traditional CNN and multi-layer perceptron (MLP) layers to incorporate SPELA. The modifications are as follows: each kernel in the convolutional layer is assigned a certain number of \textit{groups}. Each group has a nonzero number of classes. Each class belongs to exactly one group, with all classes evenly distributed across groups. Each CNN layer predicts the group assignment of an input class. For example, when the groups are (dog, cat, fish), (banana, boat, bug), and (football, airplane, phone), then the class `dog' will belong to the first group. The number of groups and classes assigned to each group varies between kernels, with randomness facilitating our performance. Each class is given a score depending on what the kernel returns. If a kernel returns the first group, all the classes corresponding to the first group get a score of one. Similarly, if a kernel returns the second group, the corresponding classes are scored, and so on. After tallying the scores, the class with the highest cumulative score is selected as the CNN layer’s output. This particular distribution of groups and classes in groups is random across kernels but is consistent across layers, one of the restrictions of our method. After obtaining the output of a particular CNN kernel, this slice of 2D data is flattened and projected down to a smaller dimension using a simple MLP. The classification is performed in the MLP precisely in a typical SPELA MLP setup. The data is pushed through the network after the classification. We keep the MLP as tiny as possible to mitigate learning in this perpendicular direction and focus more on training the CNN layer. Algorithm \ref{alg:train_cnn_layer_spela} details the layer-wise training procedure for CNNs.


\begin{algorithm}[!htb]
\caption{\textbf{Training and Inference from CNN $i^{th}$ layer with SPELA}}\label{alg:train_cnn_layer_spela}
\begin{algorithmic}[1]
\State \textbf{Given:} $\kappa$ number of classes, $C = \{1, 2, \dots, \kappa\}$, $n_i$ kernels and $B_i$ is conv block from previous layer
\State \textbf{Define:} $S_i = 0$ is score for class $i, \forall i \in C$
\State \textbf{Define:} m groups such that:
\begin{itemize}
    \item[$\rightarrow$] $G_i \subset C$
    \item[$\rightarrow$] $\bigcup\limits_{i=1}^{m} G_i = C$
    \item[$\rightarrow$] $G_i \cap G_j = \phi \; \forall i \neq j$
\end{itemize}

\For{$j \gets 1$ to $n_i$} \Comment{Define MLP for each kernel}
    \State $\text{MLP}_j ~ \mathcal{N}(0, 1)_{* \times d}$ 
\EndFor

\For{$j \gets 1$ to $n_i$} \Comment{Kernel + MLP operation}
    \State $o_j = \text{CNN}(B_i, k_j)$ \Comment{$k_j$ is the $j^{th}$ kernel}
    \State $o_j^{'} = \text{flatten}(o_j)$
    \State $o_j^{''} = \text{MLP}_j(o_j^{'})$
    \State $\text{loss}_j = -\text{cos}\textunderscore \text{sim}(o_j^{''})$
    \State Say the closest predicted group is $G_m$ via $\text{cos}\textunderscore \text{sim}$ loss
    \For{$c \; \text{in} \;  G_m$}
        \State $S_c = S_c + 1$
    \EndFor
\EndFor 

\For{$j \gets 1$ to $n_i$} 
    \State $w_{k_j} \leftarrow w_{k_j} - \alpha * \nabla_{k_j}(\text{loss}_k)$ \Comment{Kernel parameter update using local loss}
    \State $w_{\text{MLP}_j} \leftarrow w_{\text{MLP}_j} - \alpha * \nabla_{\text{MLP}_j}(\text{loss}_k)$ \Comment{MLP parameter update using local loss}
\EndFor 

\State \textbf{Prediction:} $\argmax_i \text{S}_i$

\end{algorithmic}
\end{algorithm}


\section{Empirical Studies}
\label{sec:results}


\subsection{How does SPELA work?}
\label{sec:spela_work}

\begin{figure}[!htb]
    \subfloat[Loss Curve w.r.t Epochs]{%
        \includegraphics[width=5.6cm]{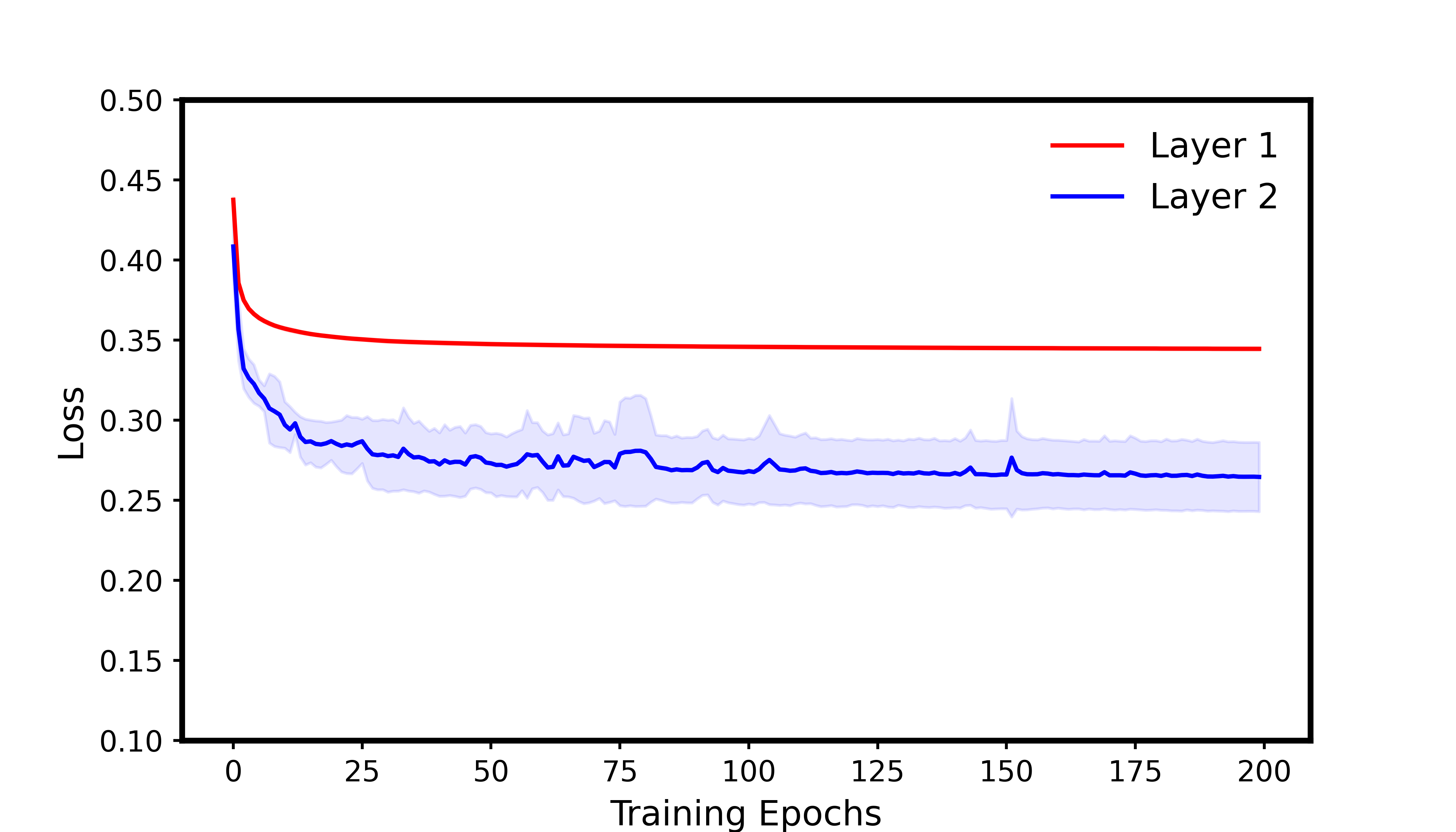}
        \label{fig:lossvsepoch_mnist}
        }%
    \hspace{1.5mm}%
    \subfloat[Accuracy w.r.t Epochs]{%
        \includegraphics[width=5.6cm]{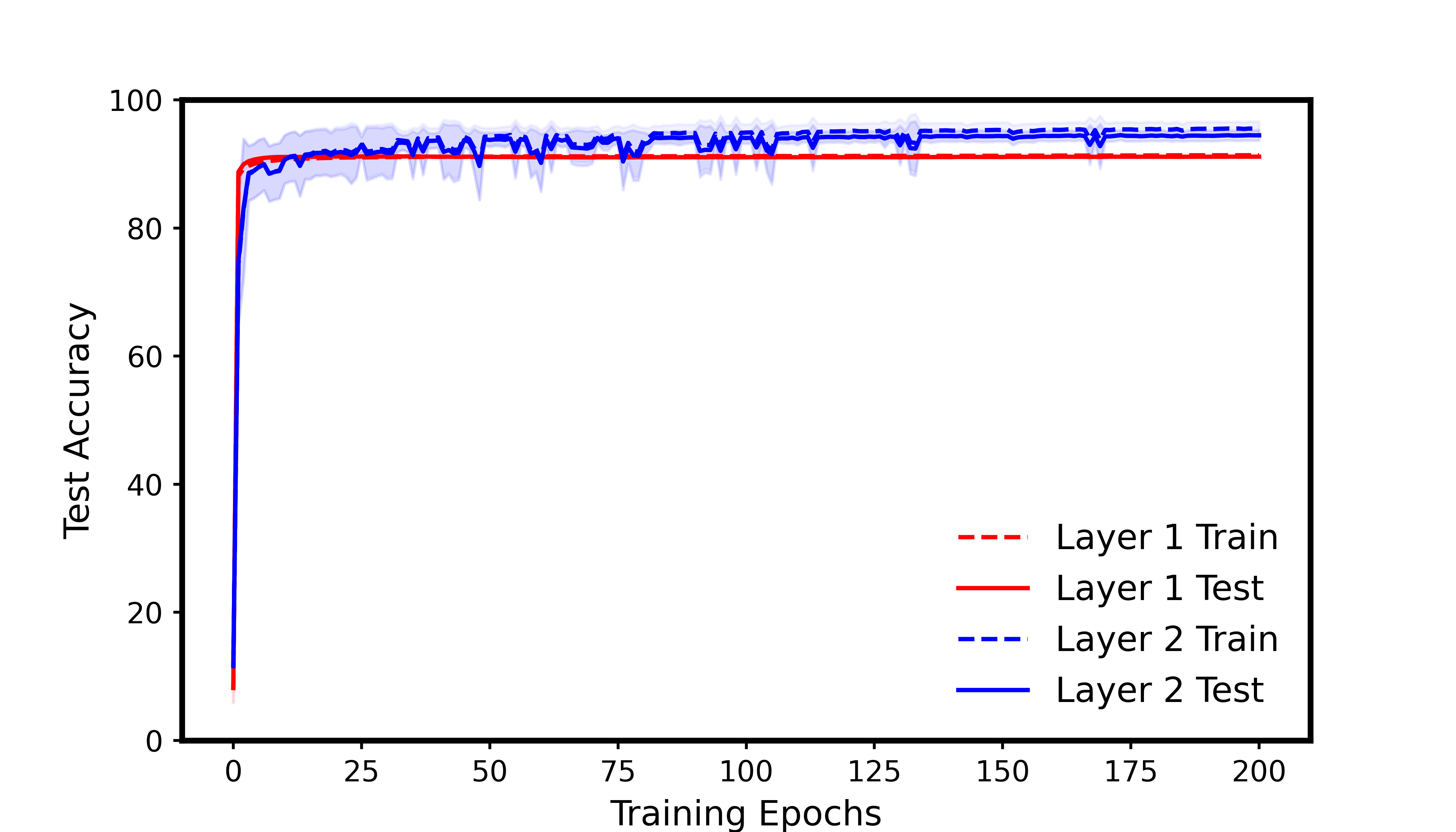}
        \label{fig:testvsepoch_mnist}
        }%
    \hspace{1.5mm}%
    \subfloat[Norm of weights w.r.t Epochs]{%
        \includegraphics[width=5.6cm]{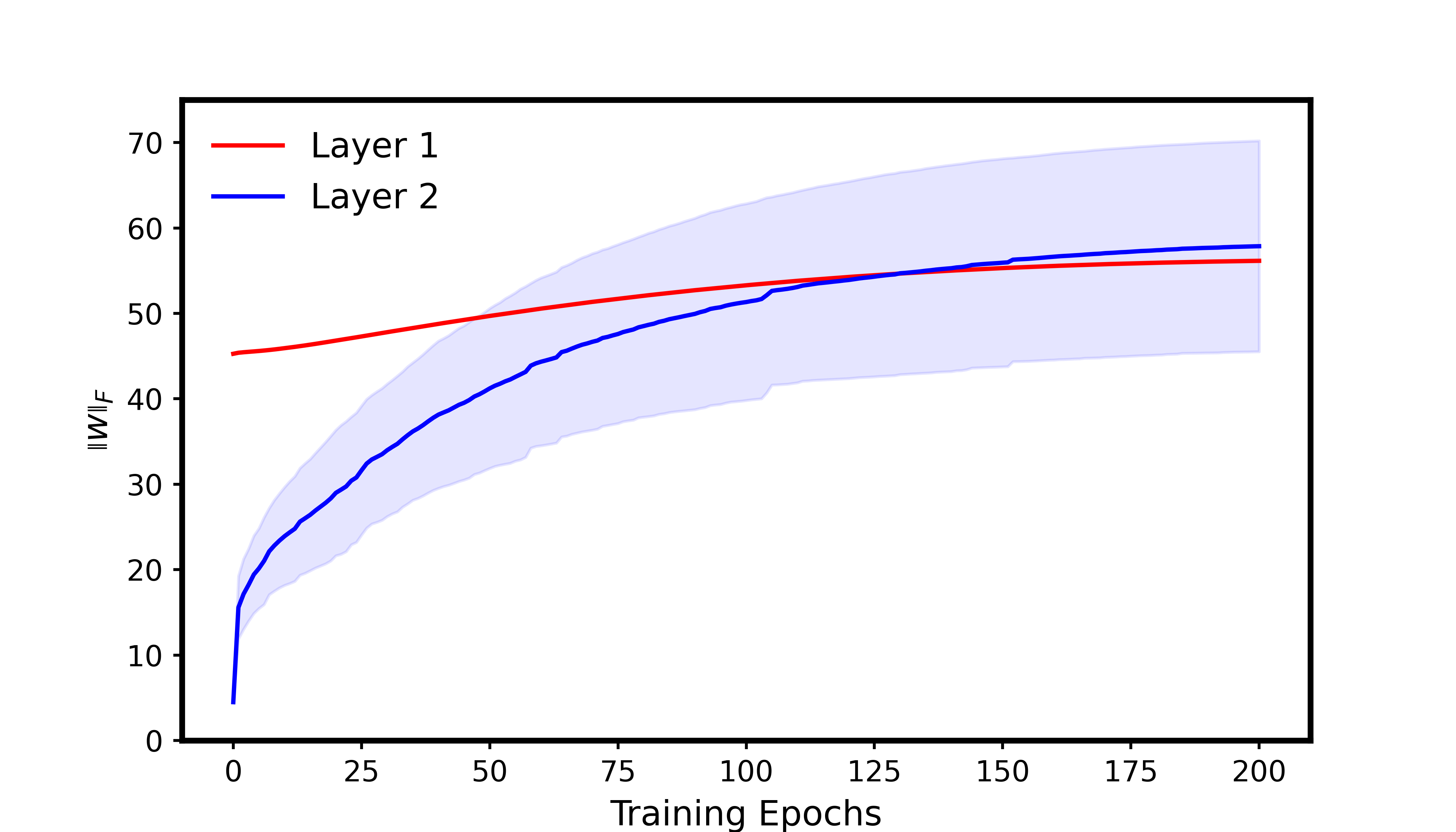}   
        \label{fig:wvsepoch_mnist}
        }%
    \caption{SPELA network behavior during learning to classify MNIST 10 digits. (a) Cosine training loss of SPELA to the progression of epochs. (b) Train and Test accuracy of both the hidden layer($1024$ neurons) and the output layer($10$ neurons) over epochs. (c) Evolution of Frobenius norm of the layer weights for SPELA with learning. The solid lines denote the mean, and the shades denote the standard deviation of five simulation runs. }
    \label{fig:perfvsepoch_mnist}
\end{figure}

We perform an in-depth analysis of SPELA's capacity. We first understand SPELA's learning dynamics on the standard MNIST 10 dataset. Following the design described in \citet{dellaferrera22a}, we evaluate the performance of a $784 \rightarrow 1024 \rightarrow10$ size SPELA network on MNIST 10. The network is trained by cosine loss defined as $\log$(2 - Cosine Similarity). Figure \ref{fig:lossvsepoch_mnist} describes both layers of SPELA's epoch-specific decreasing training loss curves. The mean loss for layer 2 ($0.26$ after 200 epochs) is always lower than the corresponding layer 1 loss ($0.34$ after 200 epochs). Figure \ref{fig:testvsepoch_mnist} describes the rise of test accuracy with the number of training epochs with SPELA. It is observed that SPELA training is effective at improving the accuracy from $11.60\%$ without learning to $94.49\%$ after $200$ epochs of training. By design, SPELA can perform predictions at all layers (except the input layer), wherein the amount of available resources can determine the number of layers. Figure \ref{fig:testvsepoch_mnist} also shows that the accuracy increases with the layer counts on MNIST 10 ($91.18\%$ accuracy for layer one vs. $94.49\%$ accuracy for layer two after $200$ epochs of training), thereby justifying the need for multiple layers to improve network performance. This also empowers SPELA with an early exit feature \citep{Scardapane_2020}, enabling easy neural network distribution across hardware platforms and improving inference. Moreover, it is observed that SPELA reaches near maximum performance very quickly. SPELA achieves $74.88\%$ performance after only $1$ epoch of training on MNIST 10. Next, we analyze the effect of SPELA training on network parameters as in Figure \ref{fig:wvsepoch_mnist}. The Frobenius norm of the layer weights demonstrates an increasing trend with learning, precisely $45.27, 4.49$ (before learning) to $56.16, 57.86$ (after $200$ epochs) for layer 1, and layer 2, respectively. 

\begin{figure}[!htb]
     \subfloat[]{%
        \includegraphics[width=8.5cm]{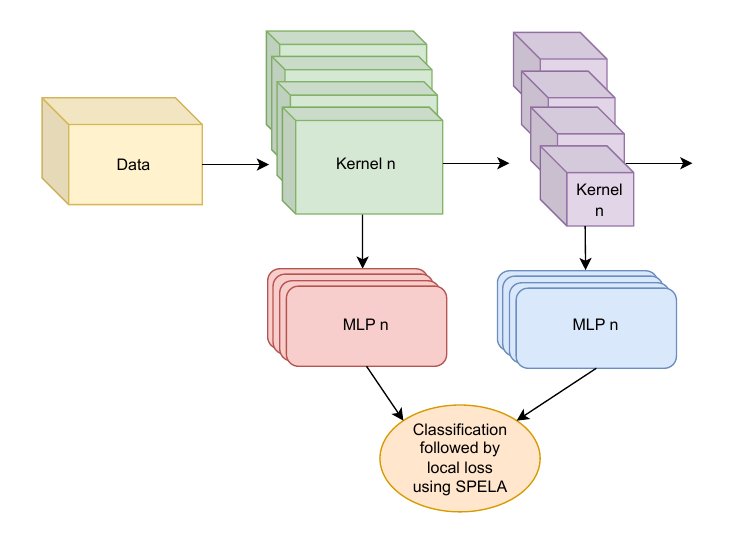}
        \label{fig:cnn_net_arch}%
        }%
    \hspace{6mm}
    \subfloat[Before Learning]{%
        \raisebox{18mm}{ \includegraphics[width=3.25cm]{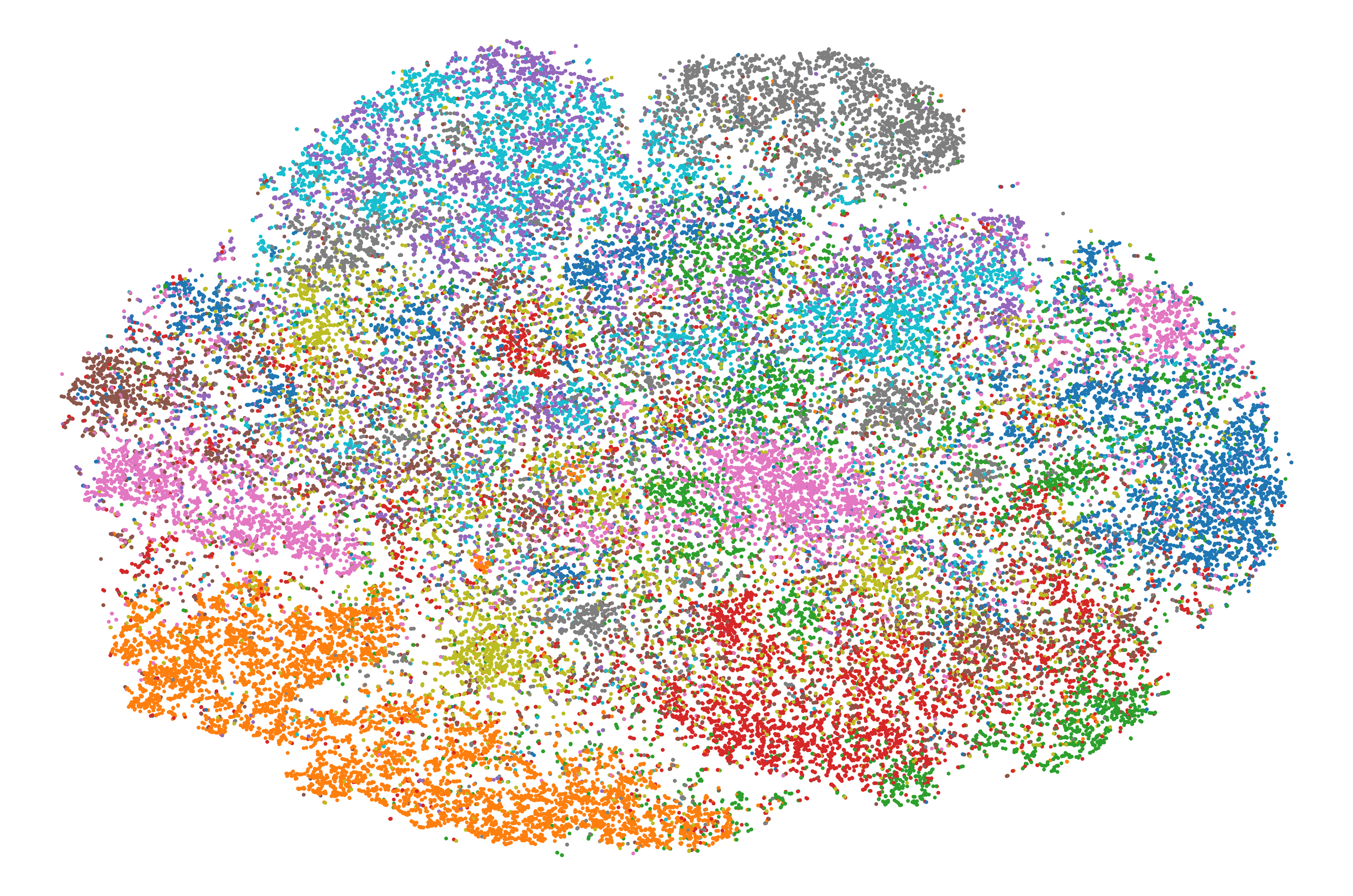}}
        \label{fig:tsne_nl}%
        }%
    \subfloat[After Learning]{%
         \raisebox{18mm}{\includegraphics[width=3.25cm]{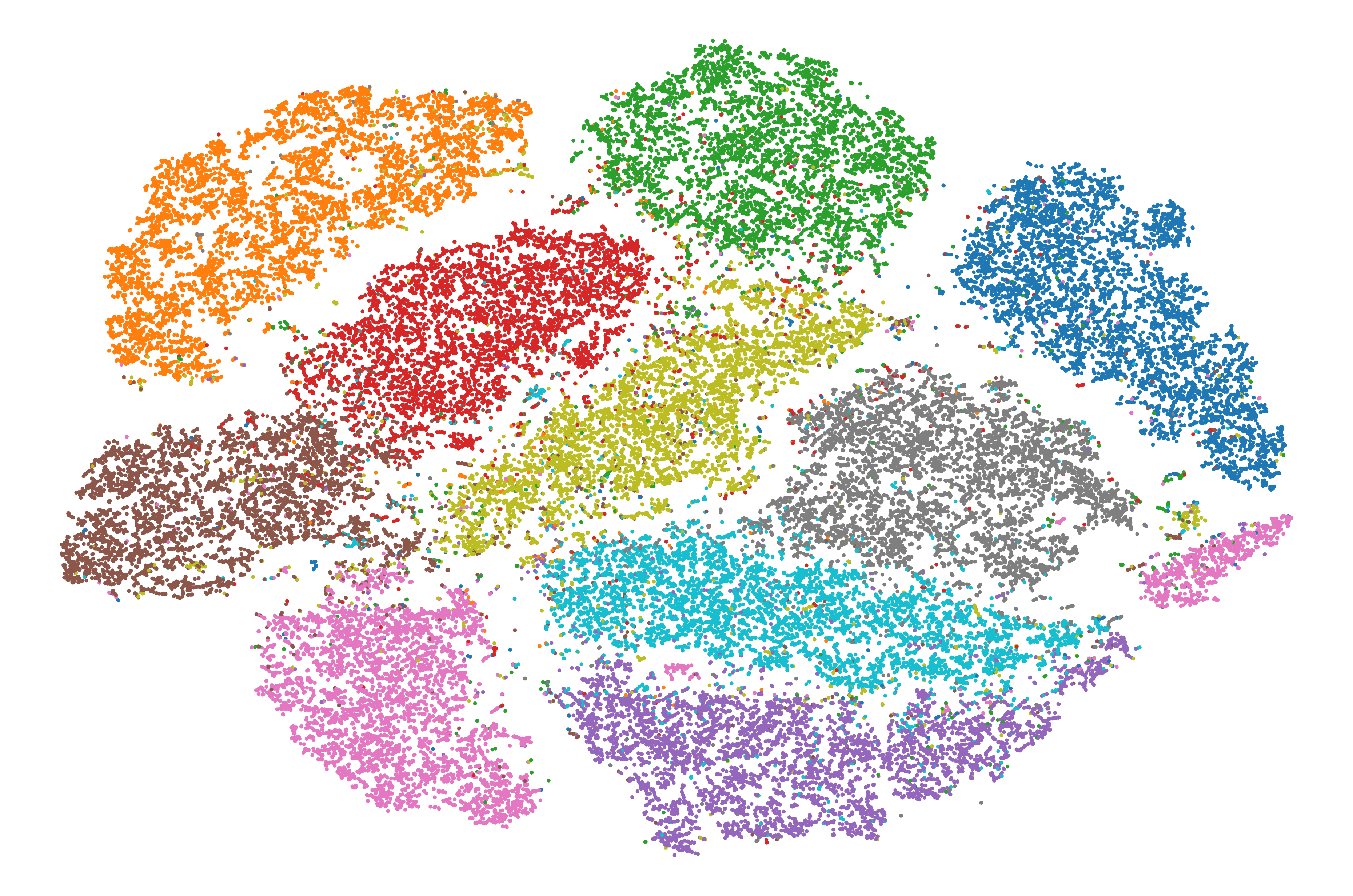}}
        \label{fig:tsne_l}
        }%
    \caption{(a) Schematic diagram of SPELA Convolutional Neural Network. (b), (c) Two-dimensional t-SNE embeddings of the output layer($10$ neurons) of a SPELA MLP network trained on MNIST 10 (b) Before any training, (c) After training for $200$ epochs. The colors corresponded to different digits of MNIST 10.}
    \label{fig:t-sne spela}
\end{figure}

Figure \ref{fig:tsne_nl} establishes the representation learning capabilities of SPELA. A t-SNE embedding analysis of the output layer representation ($10$ neurons) before learning in Figure \ref{fig:tsne_nl} exhibits a high degree of overlap of the MNIST 10 digit classes. However, after training with SPELA for $200$ epochs of the network Figure \ref{fig:tsne_l}, we observe $10$ distinct clusters corresponding to MNIST classes in the t-SNE embeddings of the output layer representation.

SPELA can predict directly by comparing the layer output with the symmetric class embeddings. However, we parallelize this comparison using a classifier head with weights determined by the class embeddings. This design can use either a cosine loss (SPELA) or a cross-entropy loss (SPELA\_CH) for layer training. We compare SPELA's performance wherever possible with existing work.
Table \ref{tab:std_split_acc_1_mlp-1} describes the key comparison results on MNIST 10, KMNIST 10, and FMNIST datasets (we selected these datasets as they are well suited for a multilayer perceptron classification task). Table \ref{tab:std_split_acc_1_mlp-2} similarly describes the performance of SPELA on reasonably complex datasets such as CIFAR 10, CIFAR 100, and SVHN 10. Notably, across both tables, SPELA is the only training method that operates with a single forward pass; in contrast, backpropagation (BP), feedback alignment (FA), direct random target projection (DRTP), and PEPITA require either a forward pass followed by a backward pass, or two consecutive forward passes.

To ensure a fair comparison, we evaluate SPELA using two configurations: (i) a baseline matching the setup from previous work \citep{dellaferrera22a} (SPELA\_CH\_A and SPELA\_A), and (ii) an optimized configuration (SPELA\_CH\_B and SPELA\_B). Full experimental details are provided in Tables \ref{tab:spela_mlp_exp_details_A} and \ref{tab:spela_mlp_exp_details_B}.

On the MNIST 10 dataset, SPELA achieves an accuracy of only $3.91\%$ lower without any dropout than an equivalent backpropagation-trained network ($94.49\%$ vs. $98.40\%$). Interestingly, a $784 \rightarrow 1000$ SPELA achieves a $91.47\%$ performance for the same training. Using the same network configuration, SPELA achieves $77.13\%$ on KMNIST-10 and $85.08\%$ on Fashion MNIST. Additionally, we observe that using cross-entropy loss (SPELA\_CH) slightly improves performance on KMNIST-10 and Fashion MNIST compared to cosine loss (77.13\% vs. 76.36\%, and 85.08\% vs. 85.00\%, respectively). Conversely, on MNIST-10, cosine loss yields marginally better results (94.49\% vs. 94.14\%). Finally, incorporating dropout did not lead to performance improvements for SPELA across any of the datasets.

Keeping the input and output layer sizes fixed at 784 and 10, we varied the number of hidden layers (1024 neurons each) of SPELA from 1 to 9. Table \ref{tab:spela_depth} describes the results of such multilayer SPELA. We observe that a 4-hidden-layer SPELA performs best on MNIST (96.28\%). Similarly, a 7-hidden layer and a 4-hidden layer perform best on KMNIST 10 and Fashion MNIST 10 respectively (83.57\%, and 87.16\%). 

\begin{figure}[!htb]
    \subfloat[MNIST 10]{%
        \includegraphics[width=5.1cm]{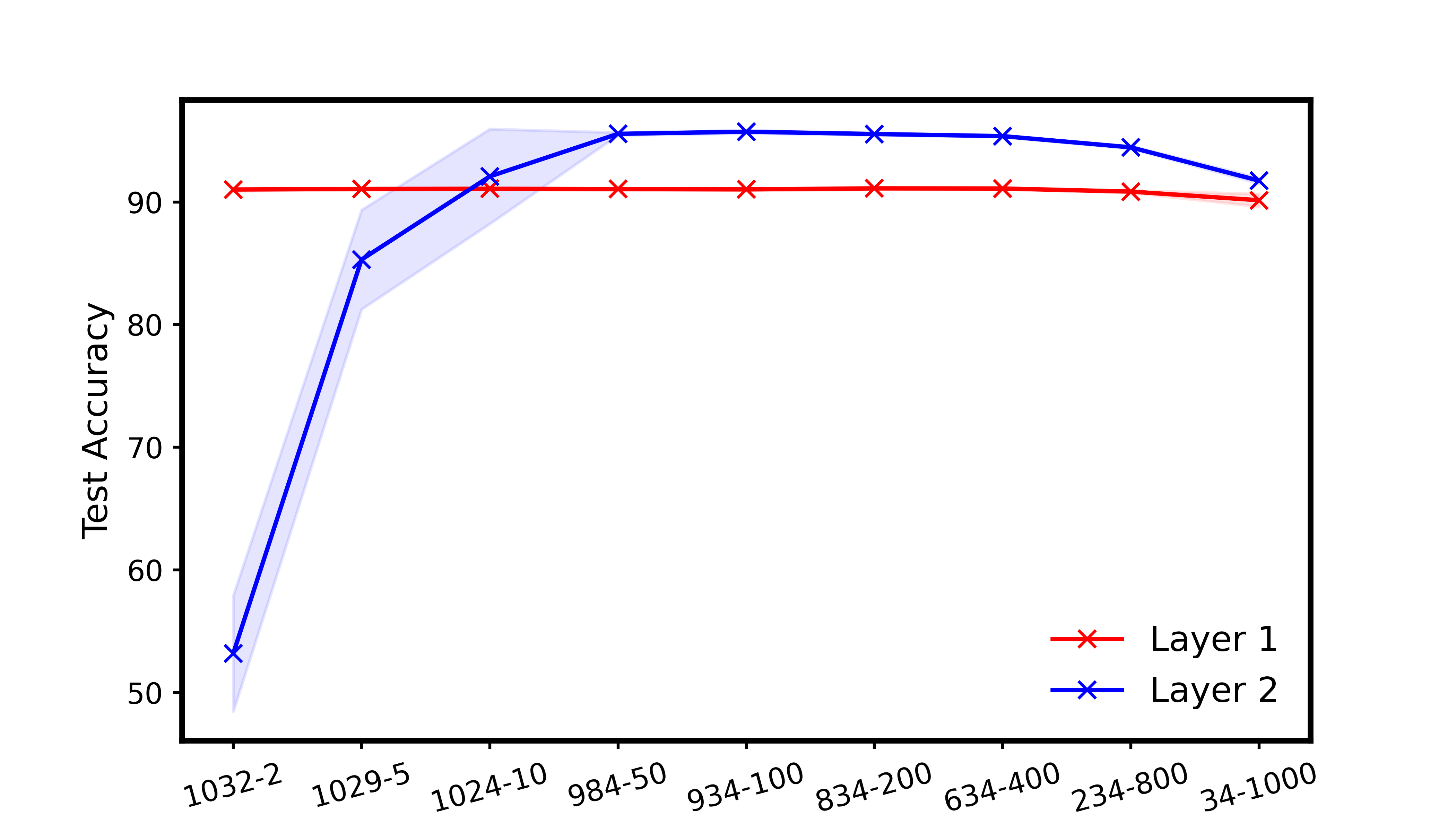} 
        }%
    \hspace{1.5mm}%
    \subfloat[KMNIST 10]{%
        \includegraphics[width=5.1cm]{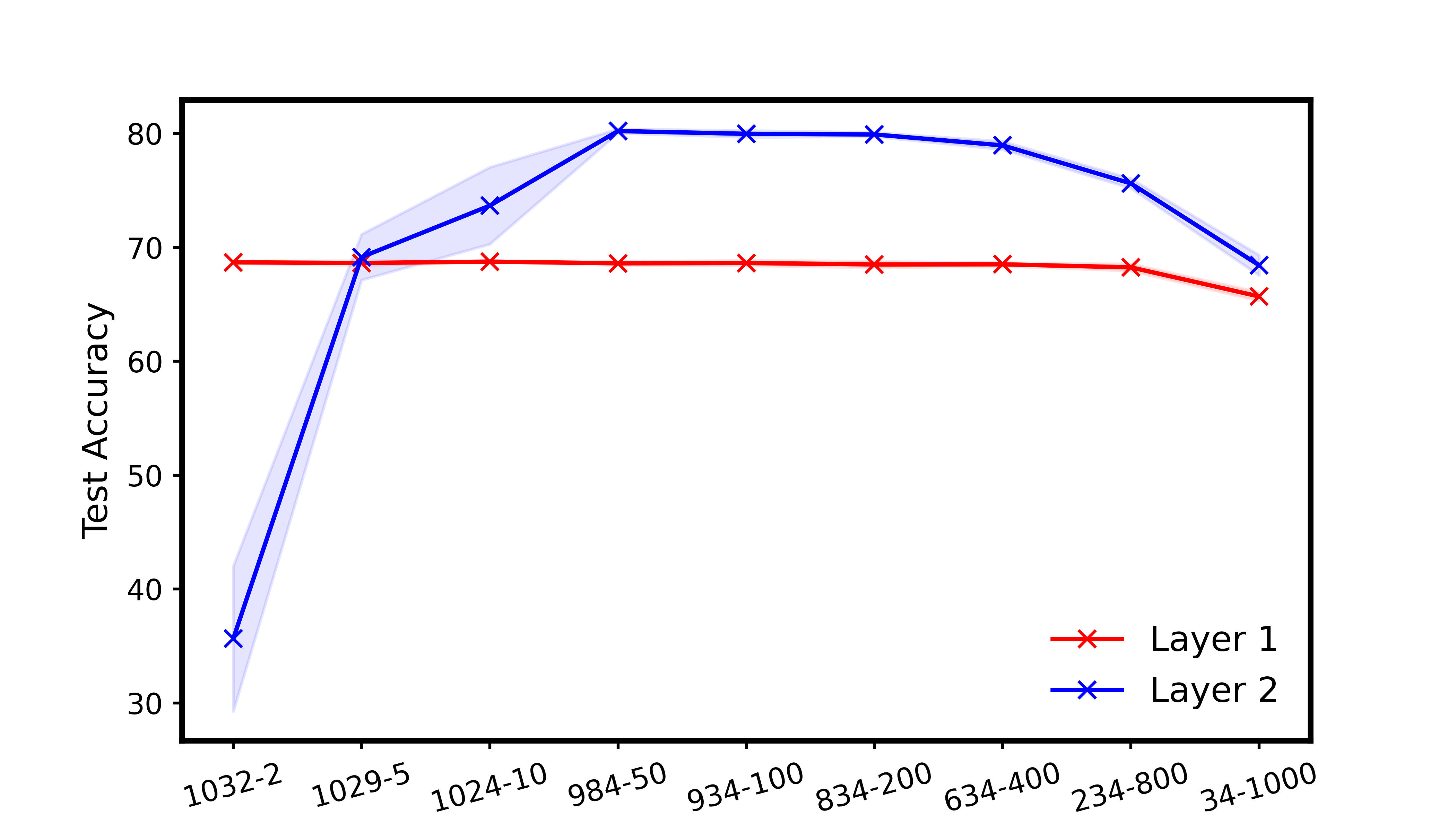} 
        }%
    \hspace{1.5mm}%
    \subfloat[Fashion MNIST 10]{%
        \includegraphics[width=5.1cm]{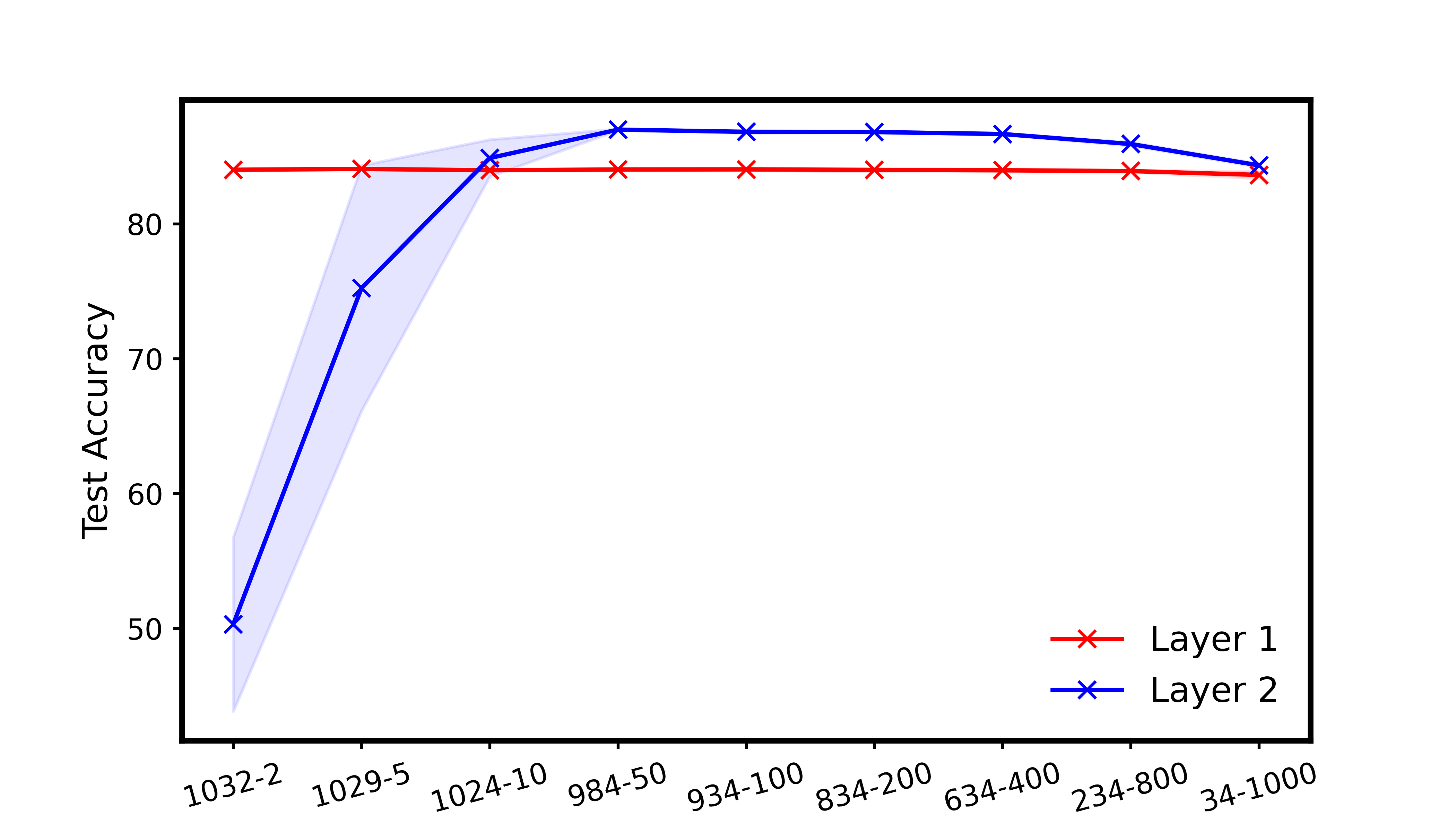} 
        }%
       
    \caption{Variation of SPELA's test performance for output layer sizes of 2, 5, 10, 50, 100, 200, 400, 800, and 1000. We keep the total number of network neurons fixed to 1034; hence, the corresponding hidden layer sizes are 1032, 1029, 984, 934, 634, 234, and 34. Throughout the experiments, the SPELA\_B configuration is used for 200 training epochs. The solid lines denote the mean, and the shades denote the standard deviation of five simulation runs.}
    \label{fig:perfalllayervary}
\end{figure}

By design, SPELA does not require a softmax layer whose dimensions are usually equal to the number of classes in the dataset. Without this constraint, we explore the optimal network configuration, keeping the total number of neurons constant ($1024+10 = 1034$). Figure \ref{fig:perfalllayervary} and Table \ref{tab:spela_outputvary} clearly show that SPELA's performance depends on output layer size. When the output layer size is 5 (and the corresponding hidden layer size is 1029), MNIST 10, KMNIST 10, and Fashion MNIST 10 performances are merely 53.19\%, 35.66\%, and 50.31\% respectively.
Conversely, a $784\rightarrow 984 \rightarrow50$ architecture, wherein the output layer size is five times that of the number of classes, achieves the best MNIST 10, KMNIST 10, Fashion MNIST 10 performances of 96.28\%, 80.23\% and 87\% respectively.
This work does not aim to compete with backpropagation (BP). Instead, our focus is on ANN training for applications where BP is computationally infeasible due to issues such as storage of forward pass activations (such as in the case of on-device learning with backpropagation \citep{tinyTL}). 

\begin{table}[!htb]
    \centering
    \begin{tabular}{|c| c| c| c| c| c|}
        \hline
        \textbf{Model} & \textbf{Architecture} & \textbf{\# Epochs} & \textbf{MNIST 10} & \textbf{\makecell{KMNIST 10}} & \textbf{\makecell{FMNIST 10}} \\
        \hline 
        BP\_A & $784\rightarrow 1024 \rightarrow 10$ & 100 & 98.56 $\pm$ 0.06 & 92.73 $\pm$ 0.11 & 88.72 $\pm$ 0.21
        \\
        \hline 
        FA & $784\rightarrow 1024 \rightarrow 10$ & 100 & 98.42 $\pm$ 0.07 & - & - 
        \\
        \hline 
        DRTP & $784\rightarrow 1024 \rightarrow 10$ & 100 & 95.10 $\pm$ 0.10 & - & -
        \\
        \hline 
        PEPITA & $784\rightarrow 1024 \rightarrow 10$ & 100 & 98.01 $\pm$ 0.09 & - & -
         \\
        \hline 
        FF & $784 \rightarrow 4\times(\rightarrow  2000)$  & 60 & 98.6 & - & -
        \\
        \hline
        SPELA\_CH\_A & $784\rightarrow1024$ & 100 & 87.02 $\pm$ 0.2 &  59.73 $\pm$ 0.23  &   73.63 $\pm$ 0.52
        \\
        \hline
        SPELA\_CH\_A & $784\rightarrow 1024 \rightarrow 10$ & 100 & 82.41 $\pm$ 5.16  &  72.52 $\pm$ 3.13  &    77.17 $\pm$ 6.11
         \\
        \hline
        SPELA\_A & $784\rightarrow1024$ & 100 & 89.55 $\pm$ 0.06 &  64.27 $\pm$ 0.19  &  79.25  $\pm$ 0.11 
         \\
        \hline
        SPELA\_A & $784\rightarrow 1024 \rightarrow 10$ & 100 & 90.27 $\pm$ 4.17 &  64.71 $\pm$ 6.96  &  72.58  $\pm$ 8.23
        \\
        \hline
        BP\_CH\_B & $784\rightarrow 1024 \rightarrow 10$ &  200 & 98.40 $\pm$  0.08 & 88.81  $\pm$ 2.03 & 91.89 $\pm$ 0.23\\
         \hline
        SPELA\_CH\_B & $784\rightarrow1024$ & 200 & 91.47 $\pm$ 0.06 & 68.7 $\pm$ 0.24 & 83.68 $\pm$ 0.04\\
          \hline
        SPELA\_CH\_B & $784\rightarrow 1024 \rightarrow 10$ & 200 & 94.14 $\pm$ 1.22 &  77.13 $\pm$ 2.23 & 85.08 $\pm$ 0.58\\
         \hline
        SPELA\_B & $784\rightarrow1024$ & 200 & 91.18 $\pm$ 0.09 & 68.7 $\pm$ 0.17 & 84.01 $\pm$ 0.10\\
          \hline
        SPELA\_B & $784\rightarrow 1024 \rightarrow 10$ & 200 & 94.49 $\pm$ 0.74 & 76.36 $\pm$ 2.79 & 85 $\pm$ 1.2\\

        \toprule
        
    \end{tabular}
    \caption{Test accuracies (mean $\pm$ standard deviation) comparison of different learning methods on MNIST 10, KMNIST 10, and FMNIST 10 (Fashion MNIST 10) datasets. The accuracies of FA, DRTP, and PEPITA are as presented in  \citet{dellaferrera22a}.  FF results are reported in \citet{hinton2022forwardforward} for a network of 4 hidden layers with 2000 neurons each. We report both hidden layer and output mean accuracies (average of five runs) of SPELA. Table  \ref{tab:spela_mlp_exp_details_A} and Table  \ref{tab:spela_mlp_exp_details_B} describe all the relevant experiment details.}
    \label{tab:std_split_acc_1_mlp-1}
\end{table}

\begin{table}[!htb]
    \centering
    \begin{tabular}{|c| c| c| c| c| c| c| c| c| c|}
        \hline
        \textbf{Dataset} & \textbf{\# 1} & \textbf{\# 2} & \textbf{\# 3} & \textbf{\# 4} & \textbf{\# 5} & \textbf{\# 6} & \textbf{\# 7} & \textbf{\# 8} & \textbf{\# 9}\\
        \hline 
        MNIST 10 &  \makecell{ 91.21\\ $\pm$ 0.11} &  \makecell{  95.7  \\ $\pm$ 0.08} &  \makecell{  96.23\\ $\pm$ 0.07} &  \makecell{96.28\\ $\pm$ 0.09} &  \makecell{ 96.22\\ $\pm$ 0.05} &  \makecell{ 96.26 \\ $\pm$ 0.08} &  \makecell{96.15 \\ $\pm$ 0.06 } &  \makecell{96.16 \\ $\pm$ 0.06 } & \makecell{ 96.14 \\ $\pm$ 0.08}

        \\
        \hline 
        KMNIST 10 &  \makecell{ 68.70\\ $\pm$  0.11} &  \makecell{ 80.32\\ $\pm$ 0.27} &  \makecell{ 83.11\\ $\pm$ 0.13} &  \makecell{83.34 \\ $\pm$ 0.06} &  \makecell{83.47 \\ $\pm$ 0.21} &  \makecell{83.48 \\ $\pm$ 0.19 } &  \makecell{ 83.57\\ $\pm$ 0.20} &  \makecell{ 83.16\\ $\pm$ 0.2} & \makecell{ 83.23\\ $\pm$ 0.22}

        \\
        \hline 
        Fashion MNIST 10 &   \makecell{83.98 \\ $\pm$ 0.07}  &  \makecell{86.81 \\ $\pm$ 0.09}  &  \makecell{87.14 \\ $\pm$ 0.08}  &  \makecell{87.16 \\ $\pm$ 0.06} & \makecell{87.16 \\ $\pm$ 0.08} & \makecell{87.12 \\ $\pm$ 0.07} &  \makecell{87.07 \\ $\pm$ 0.13} & \makecell{87 \\ $\pm$ 0.15}   &   \makecell{86.95 \\ $\pm$ 0.06}  
        \\

        \toprule
        
    \end{tabular}
    \caption{Test accuracies (mean $\pm$ standard deviation) of SPELA for varying network depth. Columns indicate the number of 1024-neuron hidden layers in the network. For instance, column \#1 denotes a $784\rightarrow 1024\rightarrow 10$ network, \#2 denotes a $784\rightarrow1024\rightarrow1024\rightarrow10$ network. Each network with SPELA\_B configuration is trained for 200 epochs and five runs.}
    \label{tab:spela_depth}
\end{table}    

\subsection{Training Memory Comparison}

\begin{figure}[!htb]
    \centering
        \includegraphics[width=10cm]{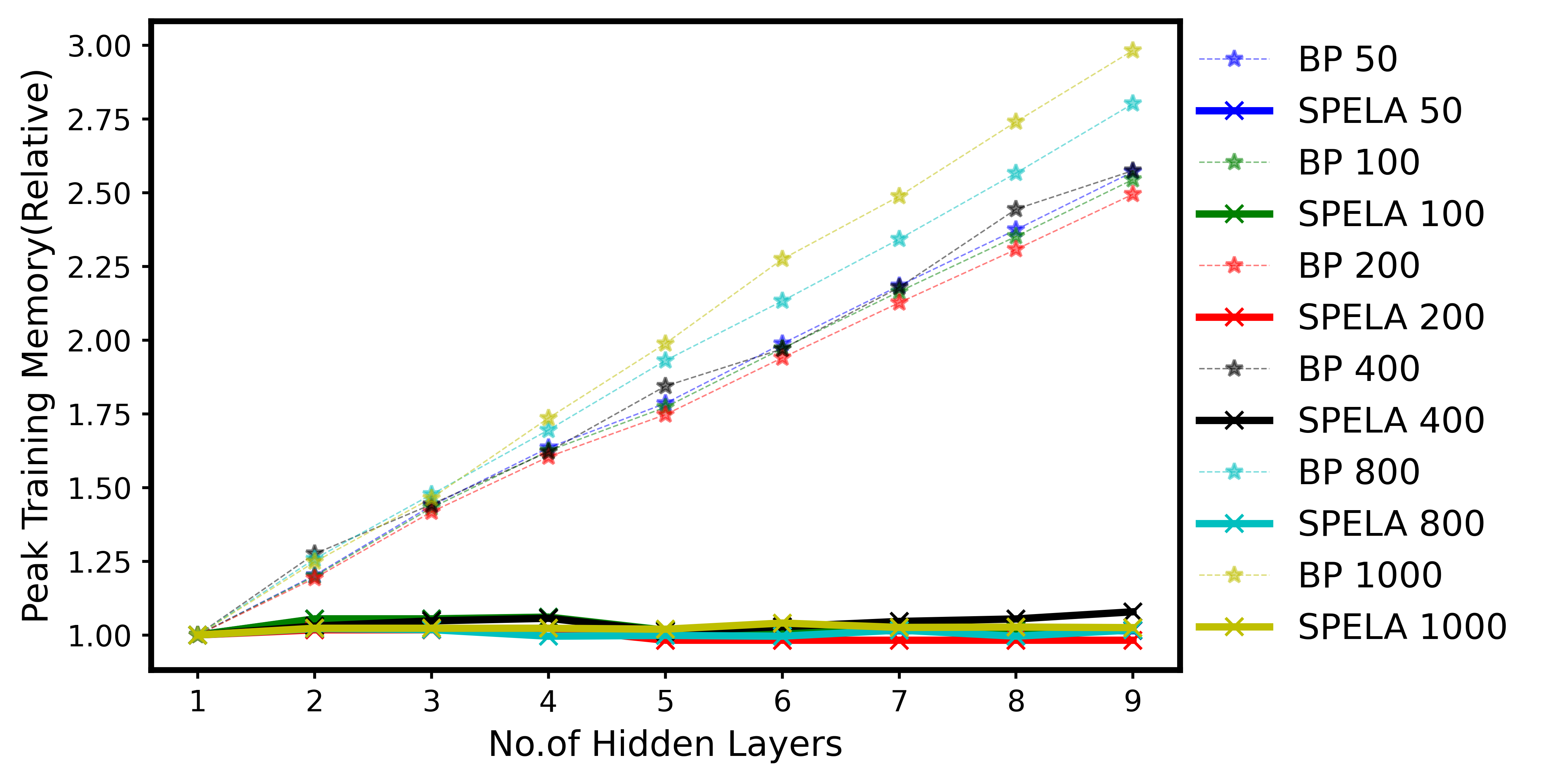}
    \caption{ Variation of relative peak training memories(average of five runs) with layer depth for SPELA and BP. We use the peak training memory occupied by a $784\rightarrow 1024\rightarrow 10$ SPELA/BP network to report the relative peak training memory. The hidden layer sizes vary from 1-9 with 1024 neurons in each hidden layer, and for batch sizes of 50, 100, 200, 400, 800, and 1000.}
  
    \label{fig:spelavsbp_mem}
\end{figure}

Next, we analyzed and compared SPELA's peak training memory consumption with an equivalent BP network. Training memory typically dominates model memory during learning; hence, we focus on peak training memory usage. Figure \ref{fig:spelavsbp_mem} presents the relative peak GPU memory consumption when the number of hidden layers varies from 1 to 9. For all batch sizes: 50, 100, 200, 400, 800, and 1000, the relative peak training memory consumed by SPELA training is almost the same, resulting in a flat curve. We use the training memory consumed by a $784 \rightarrow 1024 \rightarrow 10$ as a base for computing the relative peak training memory. Meanwhile, for a BP network, the relative peak training memory increases with the number of hidden layers and batch size. Notably, the training memory is the residue after subtracting the model memory from the total peak memory. Due to SPELA's design, such as a per-layer classifier head which uses the vector embeddings as weights, the model memory of SPELA should be very slightly higher than that of BP, which we observe in Table \ref{tab:spela_model_mem} (for instance, 38.20 MB vs. 37.83 MB for a nine-hidden-layer layer network). 

\subsection{Transfer Learning with SPELA}
\label{sec:spela_tinytl}

\begin{figure}[!htb]
  \centering
  \subfloat{\includegraphics[width=1.\linewidth]{
  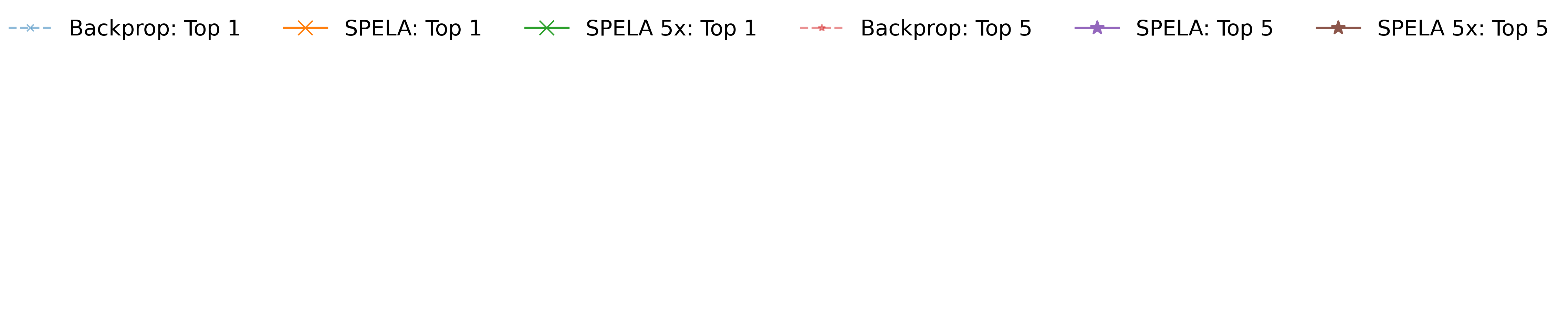}}\\
  \vspace{-3.cm}
  \subfloat[CIFAR 10 dataset] {\includegraphics[width=.33\linewidth]{
  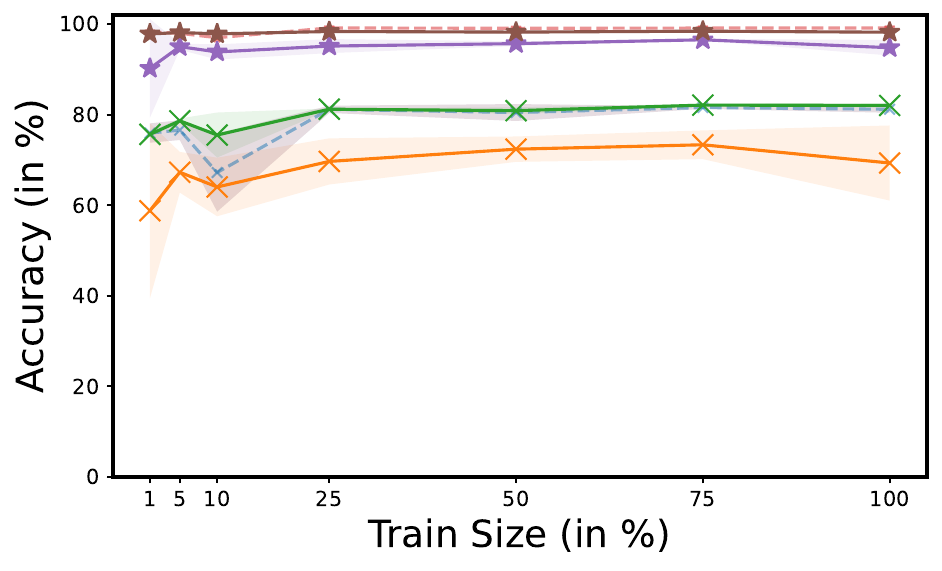}}
  \subfloat[CIFAR 100 dataset] {\includegraphics[width=.33\linewidth]{
  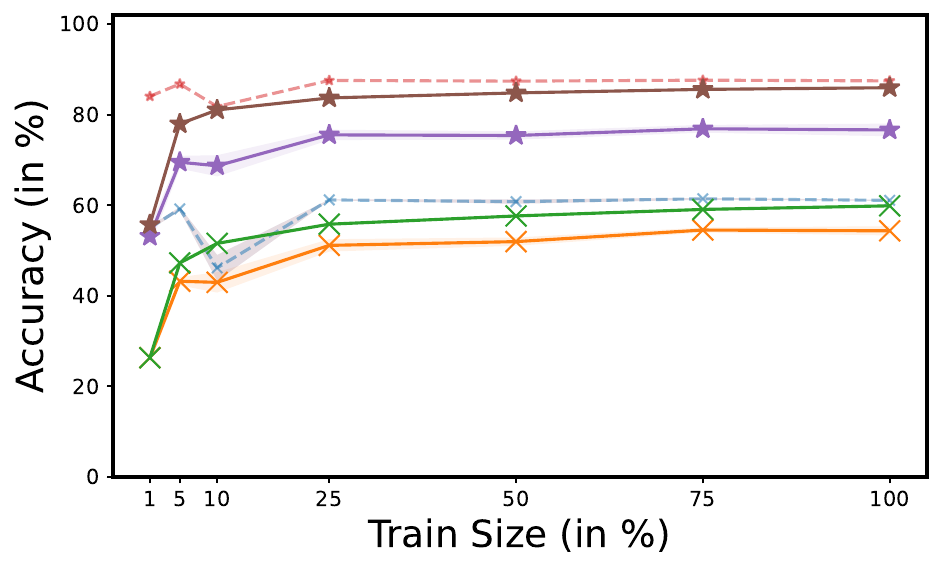}}
  \subfloat[Pets 37 dataset] {\includegraphics[width=.33\linewidth]{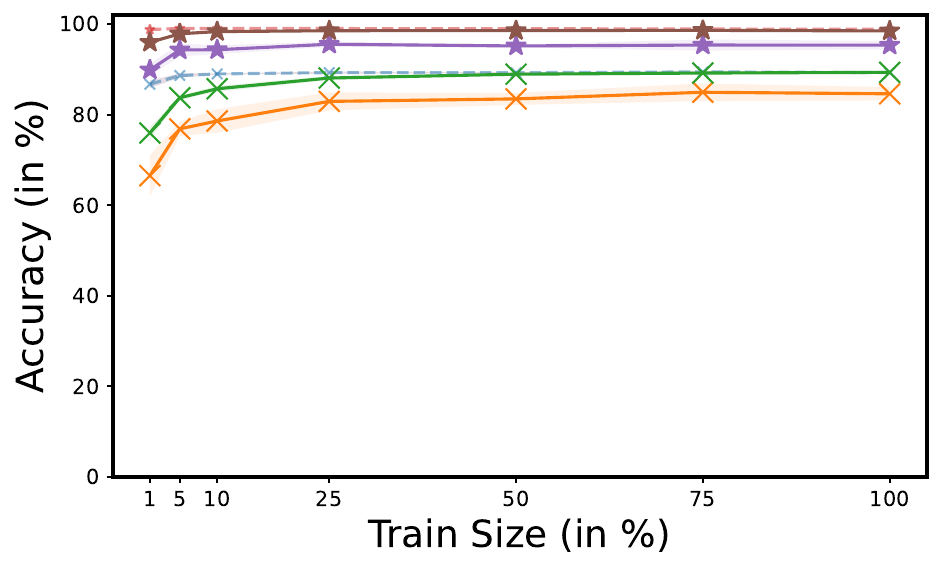}}\\
  \caption{Accuracy plots (after $200$ epochs of fine tuning) of SPELA, SPELA 5x and Backpropagation trained networks for train dataset size percentages of 1, 5, 10, 25, 50, 75, and 100 during transfer learning(keeping the test dataset fixed). The solid lines denote the mean, and the shades denote the standard deviation of five simulation runs. SPELA 5x denotes a network with a classifier layer size $5 \times$ the number of classes.}
  \label{fig:TL-acc-1}
\end{figure}

In real-world applications, on-device learning(ODL) helps mitigate the impact of phenomena such as data drift by enabling the on-device update of ML models. However, on-device learning must be performed under constraints such as memory and power for tiny ML applications. Consider a neural network (NN) with three layers: $L_{0}, L_{1}$, and $L_{2}$. Previous work on on-device learning using backpropagation shows that forward pass activation storage of such layers consumes significantly higher memory than neural network(NN) parameters \citep{tinyTL}. This assumes a network design requiring $L_{0} \rightarrow L_{1} \rightarrow L_{2}$ synapses as well as $L_{0} \leftarrow L_{1} \leftarrow L_{2}$ synapses. SPELA, on the other hand,  would need only a unidirectional synaptic connection $L_{0} \rightarrow L_{1} \rightarrow L_{2}$ - implying SPELA would need fewer connection wires. These traits should make SPELA useful in tinyML applications. In this section, we examine the behavior of SPELA in the framework of tiny transfer learning, wherein the models are pre-trained with backpropagation and optimized using SPELA. We compute the top-1 and top-5 accuracies for varying degrees of training data(keeping the test dataset fixed). We follow the canonical transfer learning approach wherein the classifier head is replaced by a layer size equivalent to the number of classes. Moreover, as observed in Figure \ref{fig:perfalllayervary}, since SPELA performs better when the output layer is $5\times$ the number of classes, we also evaluate performance on SPELA 5x. Figure \ref{fig:TL-acc-1}, \ref{fig:TL-acc-2} and Tables 
\ref{tab:tinyTL accuracies dataset CIFAR10}, \ref{tab:tinyTL accuracies dataset CIFAR100}, \ref{tab:tinyTL accuracies dataset Pets}, \ref{tab:tinyTL accuracies dataset Aircraft100}, \ref{tab:tinyTL accuracies dataset Food}, \ref{tab:tinyTL accuracies dataset Flowers}, describe the performance on the six datasets. Although the ResNet50 model is trained with backpropagation(BP) and should be the obvious training method, SPELA doesn't lag far behind on the six datasets. Moreover, similar to previous observations in Figure \ref{fig:perfalllayervary}, SPELA 5x always outperforms SPELA. For instance, SPELA 5x achieves a performance of 98.20\% (vs. 99.12\% for a backpropagation network) on CIFAR 10 dataset, 85.95\%(vs. 87.46\% for a backpropagation network) on CIFAR 100 dataset, 98.48\%(vs. 98.88\% for a backpropagation network) on Pets 37 dataset, 82.96\%(vs. 85.21\% for a backpropagation network) on Food 101 dataset.


\vspace{-0.3cm}
\subsection{Ablation Studies}
\label{sec:spela_ablation}

\paragraph{Why not use alternate distance?}

Just as we try to orient the vectors to their corresponding embedded direction for correct classification, another question arises: Instead of classifying data in terms of closeness concerning angle (\textit{cosine loss}), why not classify data in terms of closeness concerning distance (Euclidean loss)? Here, we run experiments by replacing our Cosine loss function defined by $\log$(2 - cosine similarity) with the Euclidean norm loss function.  Table \ref{tab:ablation_studies} shows that when tested on MNIST 10, KMNIST 10, and Fashion MNIST 10 datasets, SPELA with Euclidean distance performs significantly lower than SPELA with cosine distance after training of $200$ epochs ($75.37 \%$ vs. $94.41 \%$ for MNIST 10, $52.71 \%$ vs. $75.05 \%$ for KMNIST 10 and $67.83 \%$ vs. $84.12 \%$ for Fashion MNIST 10). 


\citet{cer2018universalsentenceencoder} justified that small angles have very similar cosines and proposed using angular similarity:  $1- \frac{\arccos(x)}{\pi}$, for classification. Table \ref{tab:ablation_studies} shows that angular loss defined by $\log$(2 - angular similarity) performs slightly better than cosine loss for KMNIST 10 ($75.44\%$ vs. $75.05\%$) and Fashion MNIST 10 ($85.1\%$ vs. $84.12\%$). On MNIST 10, angular loss performs almost at par with cosine loss ($94.39\%$ vs. $94.41\%$). 

\paragraph{Randomizing the vector embeddings}

Symmetrically distributing the vectors is intuitive and axiomatic, as explained in Section \ref{sec: sym emb}. However, the question remains: How much would the performance drop if we randomly choose these vectors? Accordingly, we rid ourselves of the complexity of finding a symmetric distribution and run experiments by drawing the embedded vectors from a Gaussian distribution $\mathcal{N}(\mathbf{0}, \mathbf{1})$ as well as from a uniform distribution $\mathcal{U}(\mathbf{-1}, \mathbf{1})$. Table \ref{tab:ablation_studies} indicates that classification performance slightly drops for all three datasets when vectors are drawn at random. However, this performance drop is significantly lower than Euclidean distance in SPELA (MNIST 10: $4\%$ vs. $19.54\%$, KMNIST 10: $6.65\%$ vs. $22.34\%$, Fashion MNIST 10: $0.09\%$ vs. $16.29\%$). In Appendix \ref{sec: random_vec_appendix}, we discuss this observation further. 

Except for cosine and angular similarity in all of these studies, for SPELA, the first layer classification is better than the subsequent layer in the early exit set-up. 

\paragraph{Binarization ($\pm1$) of SPELA}

\citet{g.2018the} shows that binarization does not significantly change the directions of the high-dimensional vectors. As our algorithm tries to orient the activations to a particular direction in high dimensions, it is safe to assume that the weight binarization should not significantly affect the performance. In Table \ref{tab:ablation_studies}, we show the results of our experiments involving the binarization of weights (here, we do not binarize the bias involved at each layer, only the weights to $\pm1$). Table \ref{tab:ablation_studies} shows that our assumption is experimentally proven accurate. Binarization of weights while dealing with vectors in high dimensions does not change the relative angular positions significantly; hence, the accuracy does not drop significantly either. This modification could lead to a far more efficient algorithm that cuts down on memory and energy consumption, as well as the area occupied by a chip, while not sacrificing performance. Noteworthily, on KMNIST 10 and Fashion MNIST 10 datasets, SPELA (with $\pm1$ weights) performs better than an equivalent backpropagation trained network (MNIST 10:$91.04 \%$ vs. $ 93.57\%$, KMNIST 10: $ 68.69\%$  vs. $ 67.98\%$, Fashion MNIST 10: $ 84.06\%$  vs. $ 76.38\%$). 

\begin{figure}[!htb]
    \subfloat[MNIST 10]{%
        \includegraphics[width=5.7cm]{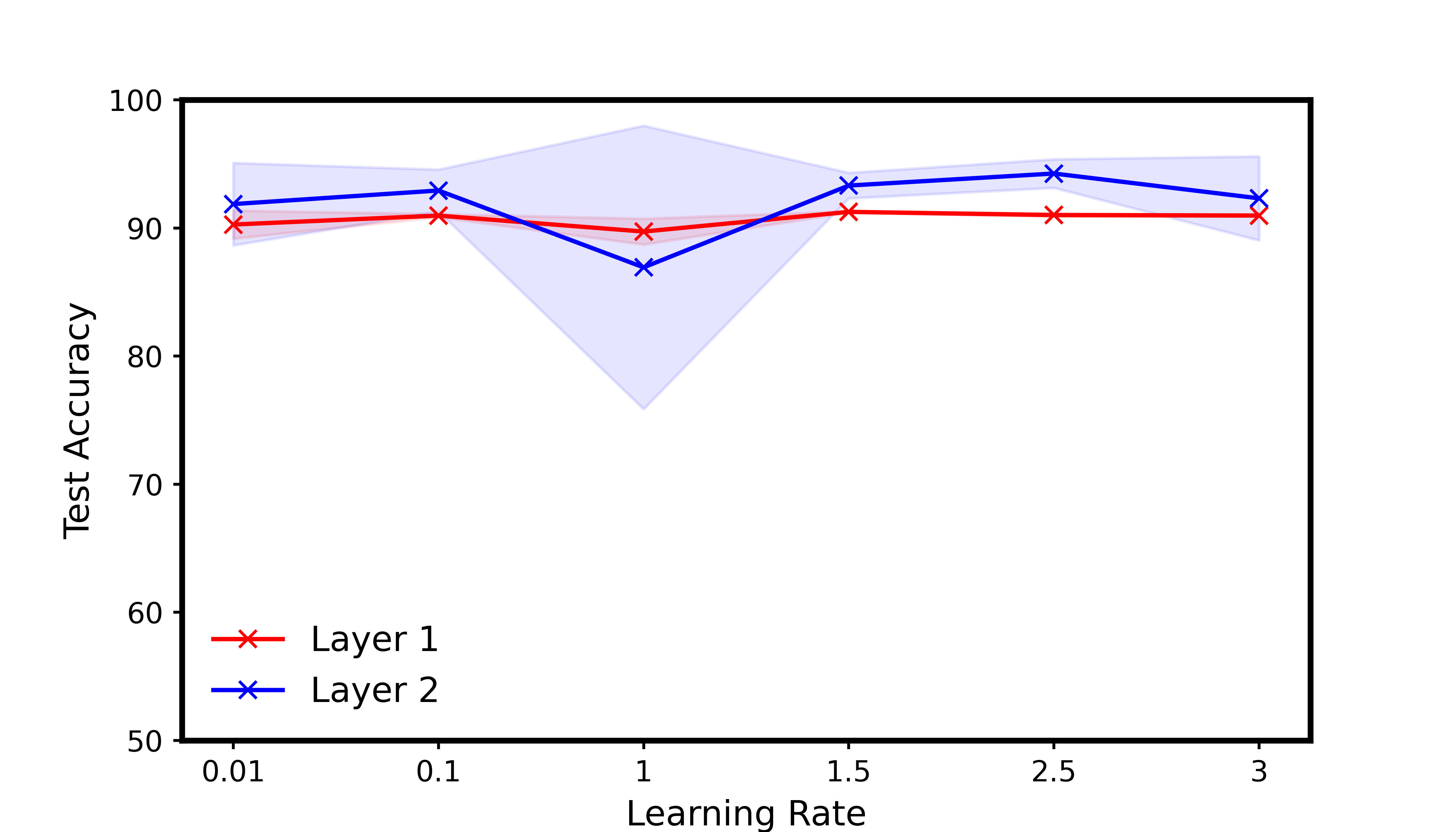} 
        }%
    \hspace{1.5mm}%
    \subfloat[KMNIST 10]{%
        \includegraphics[width=5.7cm]{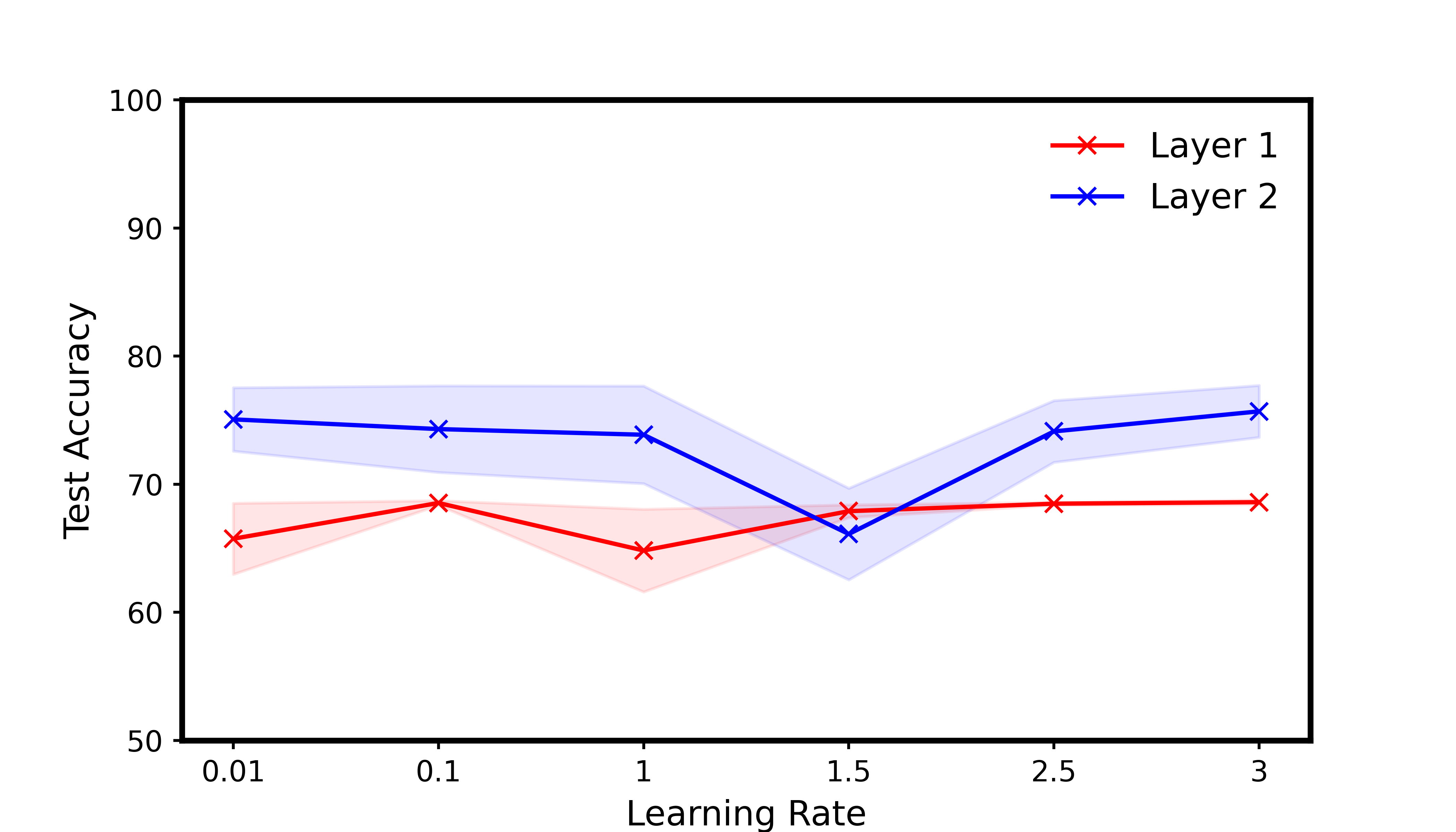} 
        }%
    \hspace{1.5mm}%
    \subfloat[Fashion MNIST 10]{%
        \includegraphics[width=5.7cm]{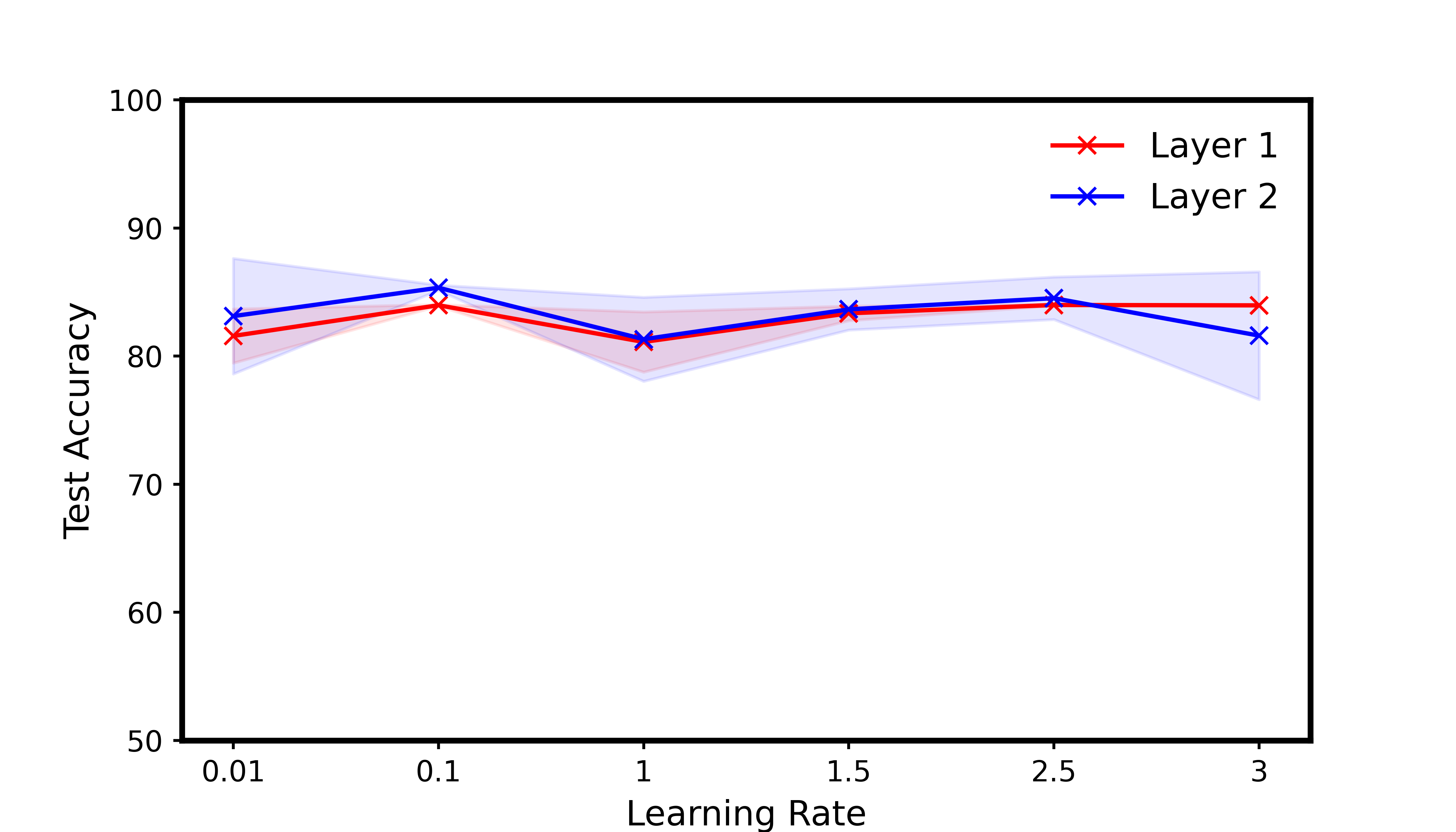} 
        }%
       
    \caption{ Variation of SPELA test accuracies (after $200$ epochs of training) for learning rates of 0.01, 0.1, 1, 1.5, 2.5, and 3. Solid lines denote mean accuracies, and shades denote standard deviation over five runs. }
    \vspace{-2.5pt}
    \label{fig:ablation_lr}
\end{figure}

\paragraph{Learning rate dependence of SPELA}
Next, we establish the learning rate dependence of SPELA's performance after $200$ training epochs. 
Accordingly, Figure \ref{fig:ablation_lr} and Table \ref{tab:spela_lr} describe the variation of test accuracies on MNIST 10, KMNIST 10, and Fashion MNIST 10 for learning rates of 0.01, 0.1, 1, 1.5, 2.5, and 3. We observe that MNIST 10 performs best ($94.25\%$) for a learning rate of 2.5, KMNIST 10 ($75.69\%$) for a learning rate of 3, and Fashion MNIST 10 ($85.34\%$) for a learning rate of 0.1. 

Table \ref{tab:spela_mlp_ablation_exp_details} provides the relevant experiment details for the above ablation studies.

\begin{table}[!htb]
    \centering
    \begin{tabular}{|c| c| c| c| c|}
        \hline
        \textbf{Model} & \textbf{Architecture} & \textbf{MNIST 10} & \textbf{\makecell{KMNIST 10}} & \textbf{\makecell{Fashion MNIST 10}} \\
        \hline 
        SPELA-Cos & $784\rightarrow1024$ &  91.09 $\pm$ 0.12 & 68.74 $\pm$ 0.18 & 84 $\pm$ 0.11\\
        \hline 
        SPELA-Cos & $784\rightarrow1024\rightarrow10$ & 94.41 $\pm$ 0.49 & 75.05 $\pm$ 3.47 & 84.12 $\pm$ 3.30\\
         \hline 
        SPELA-Arccos & $784\rightarrow1024$ & 91.03  $\pm$ 0.06 & 68.39 $\pm$ 0.12 & 83.85  $\pm$ 0.08 \\
        \hline 
        SPELA-Arccos & $784\rightarrow1024\rightarrow10$ &  94.39 $\pm$ 0.91 & 75.44 $\pm$ 2.01 &  85.1 $\pm$ 0.71 \\
        \hline 
        SPELA-Euc & $784\rightarrow1024$ &  75.37 $\pm$ 0.34 & 52.71 $\pm$ 0.28 &  67.83 $\pm$  0.77\\
        \hline 
        SPELA-Euc & $784\rightarrow1024\rightarrow10$ & 48.16  $\pm$ 8.76 & 28.16 $\pm$ 8.14 &  51.82 $\pm$ 7.18 \\
         \hline 
        SPELA-RandNorm & $784\rightarrow1024$ & 90.91 $\pm$ 0.08 & 68.29 $\pm$ 0.21 &  84.03 $\pm$ 0.09\\
         \hline 
        SPELA-RandNorm & $784\rightarrow1024\rightarrow10$ & 86.7 $\pm$ 9.76 & 68.4 $\pm$ 6.97 &  79.89 $\pm$ 6.25\\
        \hline 
         SPELA-RandUnif & $784\rightarrow1024$ & 90.27 $\pm$ 0.16 & 67.37 $\pm$ 0.34 & 83.68  $\pm$ 0.04\\
         \hline 
        SPELA-RandUnif & $784\rightarrow1024\rightarrow10$ & 84.12 $\pm$ 5.17 & 65.45 $\pm$ 4.69 &  73.08 $\pm$ 7.28\\
        \hline 
        SPELA-Bin & $784\rightarrow1024$ & 91.04 $\pm$ 0.09 & 68.69 $\pm$ 0.17 & 84.06 $\pm$ 0.04 \\
        \hline 
        SPELA-Bin & $784\rightarrow1024\rightarrow10$ & 79.39 $\pm$ 12.67 & 61.80 $\pm$ 4.02 & 74.97  $\pm$  4.09\\
        \hline 
        BP-Bin & $784\rightarrow1024\rightarrow10$ & 93.57 $\pm$ 0.83 & 67.98 $\pm$ 4.80 & 76.38  $\pm$  4.09\\
        \toprule
        
    \end{tabular}
    \caption{Ablation study results of SPELA on MNIST 10, KMNIST 10, and FashionMNIST 10 datasets after $200$ training epochs (and for five runs). SPELA-Cos implies SPELA with cosine distance, SPELA-Arccos implies SPELA with Arc cosine distance, SPELA-Euc implies SPELA with Euclidean distance, SPELA-RandNorm implies SPELA with random vectors from a normal distribution, SPELA-RandUnif implies SPELA with random vectors drawn from a uniform distribution, SPELA-Bin/BP-Bin implies with $\pm1$ weights.}
    \label{tab:ablation_studies}
\end{table}

\vspace{-0.3cm}
\subsection{SPELA Convolutional Neural Network}
\label{sec:spela_cnn}

We evaluate the performance of our SPELA convolutional neural network (CNN) on image classification tasks. Accordingly, we here enlist the performance of SPELA CNN on complex datasets such as CIFAR 10, CIFAR 100, and SVHN 10. Table \ref{tab:spela_cnn} summarizes the details of our experiments. Similar to \citet{dellaferrera22a, Poggio}, we compare the relative performance of SPELA CNN to reported results on previously proposed backpropagation alternatives (Table \ref{tab:spela_cnn}). Table \ref{tab:std_split_acc_1_mlp-2} aims to showcase that our SPELA CNN can match these performances with only local learning and early exit capabilities. Of course, the performance of the CNN will improve by adding a global error akin to backpropagation. Although SPELA doesn't require a global error signal (either through backpropagation or as a second forward pass), a one-layer SPELA CNN attains a best performance of $45.21\%$. This is $5.21\%$ lower than DRTP. Adding another layer to SPELA CNN improves performance to $56.33\%$, similar to a one-layer PEPITA. On the SVHN 10 dataset, a two-layer SPELA CNN archives a mean accuracy of $79.02\%$, which is a significant improvement over a vanilla SPELA performance on SVHN 10 ($66.89\%$, Table \ref{tab:std_split_acc_1_mlp-2}).

\begin{table}[!h]
    \centering
    \begin{center}
    \begin{tabular}{|c|c|c|c|c|c|c|}
        \hline
        \textbf{Model} & \textbf{\makecell{CIFAR 10}} & \textbf{\makecell{CIFAR 100}} & \textbf{\makecell{SVHN 10}}\\
        \hline
        BP & 64.99 $\pm$ 0.32 & 34.20 $\pm$ 0.20 & -
        \\
        \hline
        FA &  57.51 $\pm$ 0.57 & 27.15 $\pm$ 0.53 & -
        \\
        \hline
        DRTP & 50.53 $\pm$ 0.81 & 20.14 $\pm$ 0.68 & -
         \\
        \hline
        PEPITA & 56.33 $\pm$ 1.35 & 27.56 $\pm$ 0.60 & -
        \\
        \hline
        SPELA\_CH\_B CNN(1) & 45.21 $\pm$ 12.67 &  19.62 $\pm$ 5.19 & 70.10 $\pm$ 16.68
        \\
         \hline
        SPELA\_CH\_B CNN(2) & 52.4 $\pm$ 6.98 &  21.43 $\pm$ 5.94 & 79.02 $\pm$ 1.01
        \\
         \hline
        SPELA\_B CNN(1) & 44.35 $\pm$ 13.48 &  22.08 $\pm$ 4.53 & 78.14 $\pm$ 0.71
        \\
         \hline
        SPELA\_B CNN(2) & 56.59 $\pm$ 0.97 &  26.71 $\pm$ 0.79  & 76.28 $\pm$ 6.45
        \\
        
        \toprule
        
    \end{tabular}
    \end{center}
    \caption{Test accuracies (mean $\pm$ standard deviation) comparison of different CNN architectures on CIFAR 10, CIFAR 100, and SVHN 10 datasets. The accuracies of BP, FA, DRTP, and PEPITA are represented from  \citet{dellaferrera22a}. We report both hidden layer and output mean accuracies (average of five runs) of a SPELA convolutional neural network (CNN). SPELA CNN(1) and SPELA CNN(2) imply a network with a 1-layer and 2-layer CNN, respectively. BP, FA, DRTP, and PEPITA results are presented as the best-case scenario, which SPELA attempts to match under the constraint of equivalent CNN layers. }
    \label{tab:spela_cnn}
\end{table}




\vspace{-0.2cm}

\section{Discussion}

We propose SPELA as a computationally efficient learning algorithm to train neural networks. For the experiments conducted, SPELA has a feature set comprising symmetric embedded vectors, local learning, early exit, and a single forward pass for training with no storage of activations, no weight transport, and no updated weight locking. As part of it, we perform detailed experiments to benchmark SPELA against existing works. 
In addition, we analyzed its suitability for transfer learning on backpropagation-trained image recognition networks. We perform ablation studies on SPELA to justify the design choices. Finally, we extend SPELA to convolutional neural networks (CNN) and benchmark it on equivalent image recognition tasks. Analyzing theoretical complexity (lower bound) and on-device implementation shows that SPELA is more efficient than backpropagation regarding memory utilization. Hence, this work can help guide the implementation of on-device learning in tiny microcontrollers such as ARM Cortex-M devices. Overall, we believe SPELA is helpful for a wide range of ML applications, wherein we care about training and testing efficiency regarding accuracy and memory. To conclude, we view this work as a starting point for developing efficient learning algorithms that can aid in applications where backpropagation presently has limitations. In the future, we will extend SPELA to the training and inference of advanced architectures such as deep convolutional neural networks and transformers. Although SPELA is biologically inspired, it lacks some key features in biology, such as the spiking behavior of neurons. Further work is necessary to integrate such features with SPELA. 


\newpage

\bibliography{main}
\bibliographystyle{tmlr}

\newpage

\appendix

\section{Additional Methods}

\subsection{Electron simulation description}
\label{sec: sym emb}
Figuring out $N$ points on an evenly distributed circle is simple: dividing the circle into equal arcs and placing the points at their endpoints. For higher-dimensional spheres ($l_{i} \ge$ 3), the problem is non-trivial and does not admit a closed-form solution. We use a well-known method of electron approximation to simulate $N$ electrons in any $D$-dimensional space. We constrain each electron to a unit-radius ball and simulate the relative forces among electrons \citep{saff1997distributing, cohn2007universally}. At each iteration, we move each electron according to the direction and magnitude of the net force, respecting the unit-ball constraint. We define stability in terms of relative change in electrostatic potential. As the electrons shift around the ball, the electrostatic energy changes. As the electrons redistribute, their movements occur over progressively smaller distances. The relative change in electrostatic energy tends to zero with iterations. Once this relative change is less than a defined threshold, we claim to have a stable configuration and obtain an $l_{i}$-dimensional ball with a symmetric distribution of $N$ electrons (see Fig. \ref{fig:net_arch}).

\subsection{Algorithms}


\begin{algorithm}[H]
\caption{\textbf{Inference on MLP trained with SPELA}}\label{alg:alg_infer}
\begin{algorithmic}[1]
\State \textbf{Given:} An input (X) and number of layers $K$
\State \textbf{Define:} $\text{cos\textunderscore sim}(A, B) = \frac{A . B}{||A|| . ||B||}$
\State \textbf{Set:} $h_0 = x$
\For{$k \gets 1$ to $K$} \Comment{Passing data through all the layers}
    \State $h_k = \sigma_k (W_k h_{k-1} + b_k)$
\EndFor
\For{$i \gets 1$ to $N$} \Comment{N is the number of classes}
    \State $\text{S}_i = \text{cos\textunderscore sim}(h_K, \text{vecs}(i))$ \Comment{Similarity between activation vector and symmetric vectors}
\EndFor
\State \textbf{Prediction:} $\argmax_i \text{S}_i$ \Comment{Class corresponding to the maximum score is prediction}
\end{algorithmic}
\end{algorithm}

\section{Additional Results}

\subsection{How does SPELA work?}
\label{sec:spela_appendix}

\begin{table}[H]
    \centering
    \begin{center}
    \begin{tabular}{|c|c| c| c| c| c| c| c| c| }
        \hline
        \textbf{Model} & \textbf{Architecture} & \textbf{\# Epochs} & \textbf{\makecell{CIFAR 10}} & \textbf{\makecell{CIFAR 100}} & \textbf{\makecell{SVHN 10}}\\
        \hline
        BP & $3072\rightarrow1024\rightarrow10$ & 100 & 53.48 $\pm$ 0.36 & 27 $\pm$ 0.21 & 77.83 $\pm$ 0.48
        \\
        \hline
        FA & \makecell{$3072\rightarrow1024\rightarrow10$/\\$3072\rightarrow1024\rightarrow100$} & 100 & 53.82 $\pm$ 0.24 & 24.61 $\pm$ 0.28 & -
        \\
        \hline
        DRTP & \makecell{$3072\rightarrow1024\rightarrow10$/\\$3072\rightarrow1024\rightarrow100$} & 100 & 45.89 $\pm$ 0.16 & 18.32 $\pm$ 0.18 & -
         \\
        \hline
        PEPITA & \makecell{$3072\rightarrow1024\rightarrow10$/\\$3072\rightarrow1024\rightarrow100$} & 100 & 45.89 $\pm$ 0.16 & 18.32 $\pm$ 0.18 & -
        
        \\
         \hline
        SPELA\_B & $3072\rightarrow1024$ & 200 & 43.2 $\pm$ 0.18 & 19.77 $\pm$ 0.19 & 48.33 $\pm$ 5.
        \\
        \hline
        SPELA\_B & \makecell{$3072\rightarrow1024\rightarrow10$/\\$3072\rightarrow1024\rightarrow100$} & 200 & 43.41 $\pm$ 1.03 & 21.24 $\pm$ 0.12 & 66.89 $\pm$ 2.09
        \\
         \hline
        
    \end{tabular}
    \end{center}
    \caption{Test accuracies(mean $\pm$ standard deviation) comparison of different learning methods on CIFAR 10, CIFAR 100, and SVHN 10 datasets. The accuracies of FA, DRTP, and PEPITA are presented in  \citet{dellaferrera22a}. We report both hidden layer and output mean accuracies (average of five runs) of SPELA. For CIFAR 10 and SVHN 10, we use a $3072\rightarrow1000\rightarrow10$ network, and for CIFAR 100, we use a 3072-1024-100 network.}
    \label{tab:std_split_acc_1_mlp-2}
\end{table}

\begin{table}[!h]
    \centering
    \begin{tabular}{|c| c| c| c| c| c| c| c| c |c|}
        \hline
        \textbf{Dataset} & \textbf{a} & \textbf{b} & \textbf{c} & \textbf{d} & \textbf{e} & \textbf{f} & \textbf{g} & \textbf{h} & \textbf{i}\\
        \hline 
        MNIST 10 &  \makecell{ 53.19\\ $\pm$ 4.73 } &  \makecell{  85.31  \\ $\pm$ 4.05} &  \makecell{  92.09\\ $\pm$ 3.87 } &  \makecell{95.56\\ $\pm$ 0.11} &  \makecell{ 95.73\\ $\pm$ 0.13} &  \makecell{ 95.55 \\ $\pm$ 0.08} &  \makecell{95.38 \\ $\pm$ 0.07 } &  \makecell{94.46 \\ $\pm$ 0.11 } & \makecell{ 91.74 \\ $\pm$ 0.22}

        \\
        \hline 
        KMNIST 10 &  \makecell{ 35.66\\ $\pm$ 6.39 } &  \makecell{ 69.15\\ $\pm$ 2 } &  \makecell{ 73.67\\ $\pm$ 3.38 } &  \makecell{80.23 \\ $\pm$  0.18} &  \makecell{ 79.98\\ $\pm$ 0.3 } &  \makecell{ 79.92\\ $\pm$ 0.18 } &  \makecell{ 78.96\\ $\pm$ 0.36 } &  \makecell{ 75.63\\ $\pm$ 0.42 } &  \makecell{68.44 \\ $\pm$ 0.89 }  
        
        \\
        \hline 
        Fashion MNIST 10 &    \makecell{50.31 \\ $\pm$  6.45} &  \makecell{ 75.23\\ $\pm$  9.1} & \makecell{ 84.88\\ $\pm$ 1.37 } & \makecell{ 87\\ $\pm$ 0.05 } & \makecell{ 86.82\\ $\pm$ 0.09 } & \makecell{ 86.8\\ $\pm$ 0.06 } & \makecell{ 86.66\\ $\pm$ 0.06 } & \makecell{ 85.92\\ $\pm$ 0.12 }& \makecell{ 84.33\\ $\pm$  0.11}
        \\

        \toprule
        
    \end{tabular}
    \caption{Test accuracies (mean $\pm$ standard deviation) of SPELA for varying output layer depth, keeping the total neurons in the network fixed to 1034. The columns denote network configurations a: 1032-2, b: 
     1029-5, c: 1024-10, d: 984-50, e: 934-100, f: 834-200, g: 634-400, h: 234-800, i: 34-1000.  The SPELA\_B configuration is used throughout the experiments for 200 training epochs and five runs.}
    \label{tab:spela_outputvary}
\end{table}

\begin{table}[!h]
    \centering
    \begin{tabular}{|c| c| c| c| c| c| c| c| c| c|}
        \hline
        \textbf{Model} & \textbf{\# 1} & \textbf{\# 2} & \textbf{\# 3} & \textbf{\# 4} & \textbf{\# 5} & \textbf{\# 6} & \textbf{\# 7} & \textbf{\# 8} & \textbf{\# 9}\\
        \hline 
        BP &  3.26 &  7.46 &  11.65 & 15.85  & 21.03  &  25.23 & 29.43 & 33.63 & 37.83 
        \\
        \hline 
        SPELA & 3.30 & 7.54  & 11.78  & 16.02  & 21.24  &  25.48 & 29.72 & 33.96 & 38.20
      
        \\
        \toprule
        
    \end{tabular}
    \caption{Model memory(in MB) occupied by a BP model and an equivalent SPELA model for the number of hidden layers varying from 1-9 (each of size 1024 neurons). The input and output layer sizes are 784 and 10, respectively.}
    \label{tab:spela_model_mem}
\end{table}    

       

\subsection{Transfer Learning with SPELA}

\label{sec:tinytl_appendix}

\begin{figure}[!htb]
  \centering
  \subfloat{\includegraphics[width=1.\linewidth]{
  Plots/TL/TL_legend.png}}\\
  \vspace{-2.5cm}
  \subfloat[Aircraft 100 dataset] {\includegraphics[width=.33\linewidth]{
  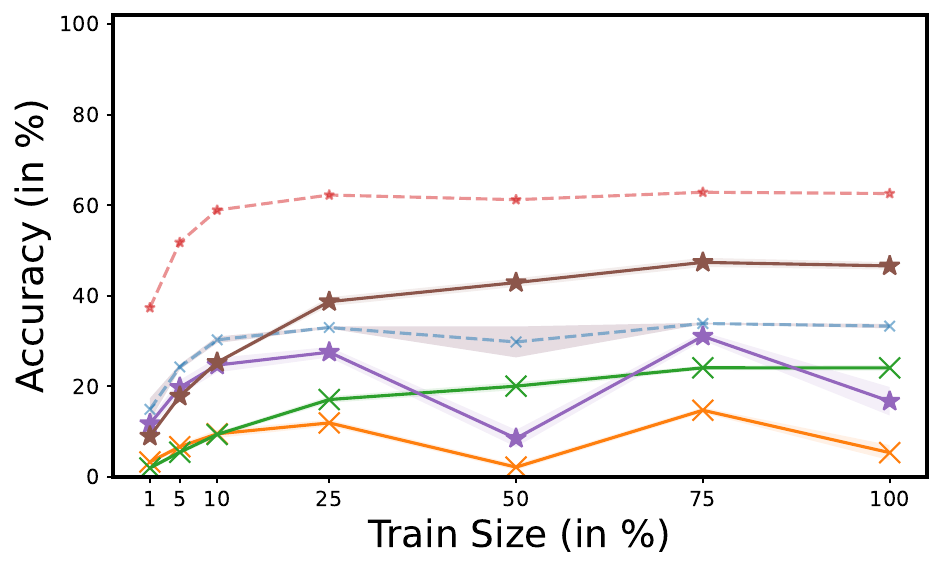}}
  \subfloat[Food 101 dataset] {\includegraphics[width=.33\linewidth]{
  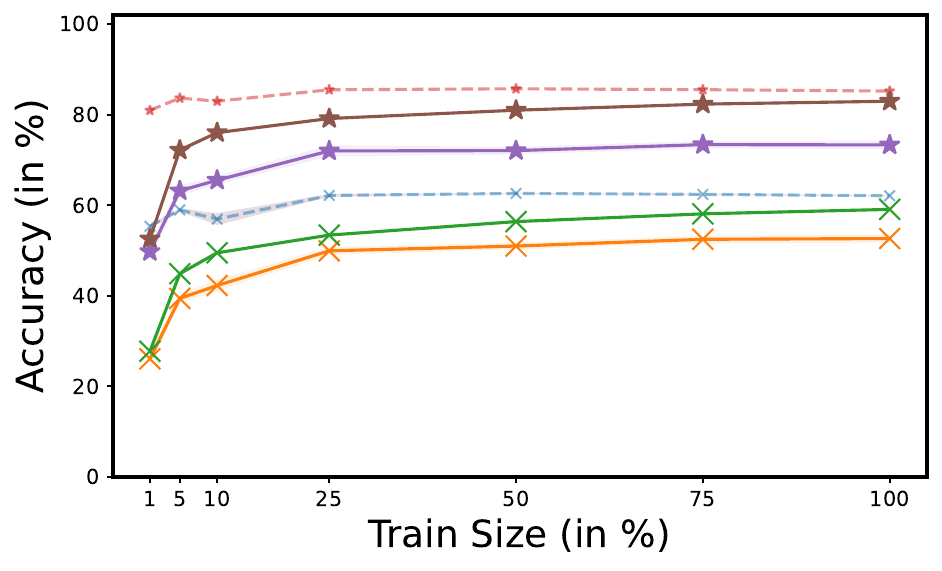}}
  \subfloat[Flowers 102 dataset] {\includegraphics[width=.33\linewidth]{
  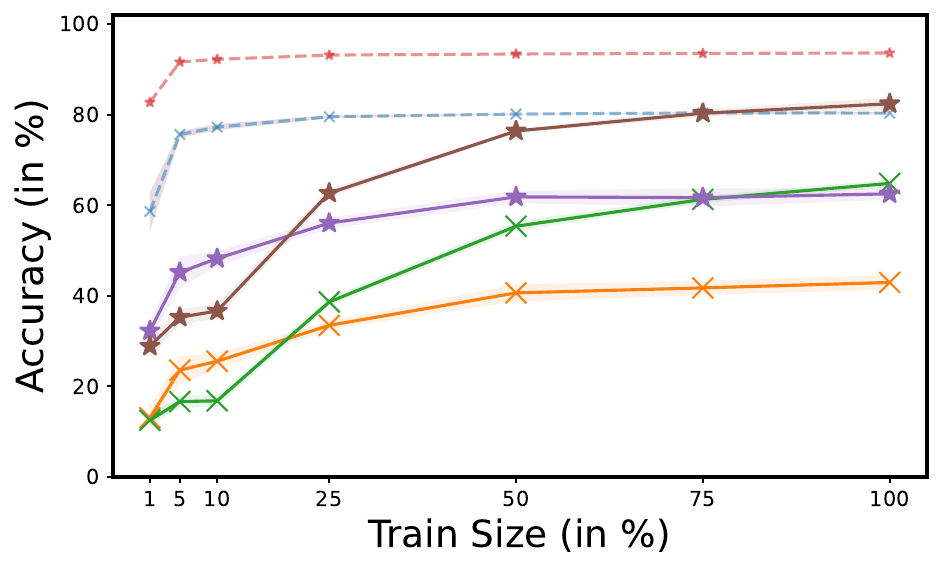}}\\
  \caption{Accuracy plots (after $200$ epochs of fine tuning) of SPELA, SPELA 5x and Backpropagation trained networks for train dataset size percentages of 1, 5, 10, 25, 50, 75, and 100 during transfer learning(keeping the test dataset fixed). The solid lines denote the mean, and the shades denote the standard deviation of five simulation runs.}
  \label{fig:TL-acc-2}
\end{figure}

In these experiments, the networks are trained for 200 epochs with a learning rate of 0.1, and analysis is done on six datasets: Aircraft 100, CIFAR 10, CIFAR 100, Flowers 102, Food 101, and Pets 37 datasets (the numbers denote the number of classes in that dataset). Note that most of these datasets have a large number of classes. Table \ref{tab:spela_mlp_transfer_learning_expdetails} describes the experimental details. A ResNet50 model pre-trained on ImageNet-1000(downloaded from PyTorch Hub) extracts features from the layer before the classifier head. These features then train a classifier head using backpropagation or SPELA.

\begin{table}[H]
    \centering
    \begin{adjustbox}{width=1.\textwidth}
    
    \begin{tabular}{|c| c| c| c| c| c| c| c| c|}
    \hline
        \textbf{Test size:} & \textbf{1} & \textbf{5} & \textbf{10} & \textbf{25} & \textbf{50} & \textbf{75} & \textbf{100}\\\hline
        
       SPELA top1 & 58.79 $\pm$ 19.33 & 67.28 $\pm$ 4.50 & 64.03 $\pm$ 6.47 & 69.67 $\pm$ 5.10 & 72.38 $\pm$ 2.80 & 73.35 $\pm$ 3.17 & 69.32 $\pm$ 8.29 \\\hline
       
        SPELA top5 & 90.23 $\pm$ 10.89 & 95.01 $\pm$ 0.69 & 93.84 $\pm$ 1.67 & 95.14 $\pm$ 1.55 & 95.68 $\pm$ 0.52 & 96.53 $\pm$ 0.20 & 94.78 $\pm$ 1.70 \\\hline

        SPELA 5x top1 & 75.64 $\pm$ 0.54 & 78.64 $\pm$ 0.53 & 75.50 $\pm$ 4.97 & 81.21 $\pm$ 0.16 & 80.88 $\pm$ 0.73 & 82.11 $\pm$ 0.32 & 82.04 $\pm$ 0.50 \\\hline

        SPELA 5x top5 & 97.81 $\pm$ 0.28 & 98.13 $\pm$ 0.24 & 97.84 $\pm$ 0.33 & 98.36 $\pm$ 0.16 & 98.22 $\pm$ 0.34 & 98.43 $\pm$ 0.05 & 98.20 $\pm$ 0.24\\\hline
        
        BP top1 & 75.93 $\pm$ 2.13 & 76.58 $\pm$ 2.18 & 67.22 $\pm$ 8.64 & 81.18 $\pm$ 0.80 & 80.47 $\pm$ 1.87 & 81.57 $\pm$ 0.27 & 81.18 $\pm$ 0.77\\\hline
        
        BP top5 & 98.25 $\pm$ 0.77 & 97.96 $\pm$ 1.70 & 96.99 $\pm$ 2.40 & 99.15 $\pm$ 0.07 & 99.06 $\pm$ 0.18 & 99.12 $\pm$ 0.12 & 99.12 $\pm$ 0.11 \\\toprule
        
    \end{tabular}
    \end{adjustbox}
    \caption{Test accuracy for CIFAR 10 dataset.}
    \label{tab:tinyTL accuracies dataset CIFAR10}
\end{table}

\begin{table}[H]
    \centering
    
    \begin{adjustbox}{width=1.\textwidth}
    \begin{tabular}{|c| c| c| c| c| c| c| c| c|}
    \hline
       \textbf{Test size:} & \textbf{1} & \textbf{5} & \textbf{10} & \textbf{25} & \textbf{50} & \textbf{75} & \textbf{100}\\\hline
      
       SPELA top1 & 26.35 $\pm$ 1.27 & 43.22 $\pm$ 0.91 & 42.97 $\pm$ 2.25 & 51.14 $\pm$ 1.37 & 51.97 $\pm$ 0.92 & 54.52 $\pm$ 0.54 & 54.35 $\pm$ 1.03\\\hline
        
        SPELA top5 & 53.14 $\pm$ 2.19 & 69.50 $\pm$ 1.41 & 68.71 $\pm$ 2.34 & 75.53 $\pm$ 1.15 & 75.40 $\pm$ 0.91 & 76.88 $\pm$ 0.85 & 76.62 $\pm$ 1.40 \\\hline

        SPELA 5x top1 & 26.33 $\pm$ 0.96 & 47.26 $\pm$ 0.30 & 51.58 $\pm$ 0.26 & 55.81 $\pm$ 0.09 & 57.63 $\pm$ 0.19 & 59.07 $\pm$ 0.30 & 59.87 $\pm$ 0.26 \\\hline

        SPELA 5x top5 & 55.74 $\pm$ 0.97 & 77.97 $\pm$ 0.31 & 81.04 $\pm$ 0.27 & 83.69 $\pm$ 0.12 & 84.80 $\pm$ 0.18 & 85.57 $\pm$ 0.06 & 85.95 $\pm$ 0.30\\\hline
        
        BP top1 & 55.00 $\pm$ 0.43 & 59.19 $\pm$ 0.40 & 46.16 $\pm$ 2.86 & 61.18 $\pm$ 0.09 & 60.79 $\pm$ 0.48 & 61.42 $\pm$ 0.20 & 61.08 $\pm$ 0.13\\\hline
        
        BP top5 & 84.05 $\pm$ 0.32 & 86.77 $\pm$ 0.25 & 81.79 $\pm$ 1.24 & 87.55 $\pm$ 0.09 & 87.40 $\pm$ 0.20 & 87.61 $\pm$ 0.07 & 87.46 $\pm$ 0.04\\\toprule
    \end{tabular}
    \end{adjustbox}
    \caption{Test accuracy for CIFAR 100 dataset.}
    \label{tab:tinyTL accuracies dataset CIFAR100}
\end{table}

\begin{table}[H]
    \centering
    \begin{adjustbox}{width=1.\textwidth}
   
    \begin{tabular}{|c| c| c| c| c| c| c| c| c|}
    \hline
       \textbf{Test size:} & \textbf{1} & \textbf{5} & \textbf{10} & \textbf{25} & \textbf{50} & \textbf{75} & \textbf{100}\\\hline
      
        SPELA top1 & 66.54 $\pm$ 4.52 & 76.86 $\pm$ 1.57 & 78.60 $\pm$ 2.61 & 82.93 $\pm$ 1.99 & 83.48 $\pm$ 1.33 & 84.94 $\pm$ 1.96 & 84.60 $\pm$ 1.46\\\hline
        SPELA top5 & 89.89 $\pm$ 1.34 & 94.28 $\pm$ 1.41 & 94.33 $\pm$ 0.98 & 95.55 $\pm$ 0.57 & 95.16 $\pm$ 0.34 & 95.37 $\pm$ 1.24 & 95.34 $\pm$ 0.91\\\hline
        SPELA 5x top1 & 75.96 $\pm$ 1.34 & 83.73 $\pm$ 0.37 & 85.71 $\pm$ 0.66 & 88.08 $\pm$ 0.42 & 88.93 $\pm$ 0.11 & 89.19 $\pm$ 0.29 & 89.34 $\pm$ 0.20\\\hline
        SPELA 5x top5 & 95.97 $\pm$ 0.45 & 97.85 $\pm$ 0.23 & 98.34 $\pm$ 0.21 & 98.53 $\pm$ 0.07 & 98.57 $\pm$ 0.06 & 98.61 $\pm$ 0.09 & 98.48 $\pm$ 0.17\\\hline
        BP top1 & 86.71 $\pm$ 0.82 & 88.64 $\pm$ 0.44 & 88.99 $\pm$ 0.16 & 89.28 $\pm$ 0.23 & 89.26 $\pm$ 0.08 & 89.44 $\pm$ 0.25 & 89.29 $\pm$ 0.10\\\hline
        BP top5 & 98.79 $\pm$ 0.11 & 99.03 $\pm$ 0.06 & 99.02 $\pm$ 0.08 & 99.01 $\pm$ 0.09 & 98.99 $\pm$ 0.04 & 98.90 $\pm$ 0.05 & 98.88 $\pm$ 0.04\\\toprule
    \end{tabular}
     \end{adjustbox}
    \caption{Test accuracy for Pets 37 dataset.}
    \label{tab:tinyTL accuracies dataset Pets}
\end{table}

\begin{table}[H]
    \centering
    \begin{adjustbox}{width=1.\textwidth}
  
    \begin{tabular}{|c| c| c| c| c| c| c| c|}
    \hline
       \textbf{Test size:} & \textbf{1} & \textbf{5} & \textbf{10} & \textbf{25} & \textbf{50} & \textbf{75} & \textbf{100}\\\hline
      
        SPELA top1 & 3.32 $\pm$ 0.43 & 6.68 $\pm$ 0.54 & 9.53 $\pm$ 0.92 & 11.94 $\pm$ 0.70 & 2.18 $\pm$ 0.64 & 14.74 $\pm$ 0.60 & 5.38 $\pm$ 1.67\\\hline
        SPELA top5 & 11.76 $\pm$ 1.13 & 19.77 $\pm$ 1.63 & 24.70 $\pm$ 1.60 & 27.55 $\pm$ 1.15 & 8.51 $\pm$ 1.87 & 31.08 $\pm$ 0.94 & 16.72 $\pm$ 3.14\\\hline
        SPELA 5x top1 & 1.97 $\pm$ 0.41 & 5.47 $\pm$ 0.53 & 9.39 $\pm$ 0.49 & 17.10 $\pm$ 0.72 & 20.06 $\pm$ 0.87 & 24.11 $\pm$ 0.45 & 24.09 $\pm$ 0.64 \\\hline
        SPELA 5x top5 & 9.00 $\pm$ 0.57 & 17.83 $\pm$ 0.67 & 25.31 $\pm$ 0.33 & 38.72 $\pm$ 0.98 & 42.92 $\pm$ 0.94 & 47.39 $\pm$ 1.01 & 46.61 $\pm$ 0.83\\\hline
        
        BP top1 & 14.94 $\pm$ 2.46 & 24.29 $\pm$ 0.56 & 30.29 $\pm$ 0.72 & 33.01 $\pm$ 0.25 & 29.82 $\pm$ 3.41 & 33.89 $\pm$ 0.16 & 33.32 $\pm$ 0.49\\\hline
        BP top5 & 37.38 $\pm$ 1.94 & 51.80 $\pm$ 0.74 & 58.93 $\pm$ 0.43 & 62.26 $\pm$ 0.20 & 61.22 $\pm$ 1.28 & 62.87 $\pm$ 0.35 & 62.59 $\pm$ 0.41\\\toprule
    \end{tabular}
     \end{adjustbox}
    \caption{Test accuracy for Aircraft 100 dataset.}
    \label{tab:tinyTL accuracies dataset Aircraft100}
\end{table}

\begin{table}[H]
    \centering
  
    \begin{adjustbox}{width=1.\textwidth}
    \begin{tabular}{|c| c| c| c| c| c| c| c| c|}
    \hline
      \textbf{Test size:} & \textbf{1} & \textbf{5} & \textbf{10} & \textbf{25} & \textbf{50} & \textbf{75} & \textbf{100}\\\hline
        
       SPELA top1 & 26.09 $\pm$ 1.72 & 39.43 $\pm$ 0.91 & 42.28 $\pm$ 1.24 & 49.93 $\pm$ 0.89 & 50.99 $\pm$ 0.73 & 52.48 $\pm$ 0.77 & 52.65 $\pm$ 0.96\\\hline
        SPELA top5 & 49.77 $\pm$ 1.61 & 63.19 $\pm$ 1.15 & 65.53 $\pm$ 0.93 & 71.99 $\pm$ 1.25 & 72.06 $\pm$ 0.82 & 73.38 $\pm$ 0.83 & 73.32 $\pm$ 0.91\\\hline
        SPELA 5x top1 & 27.77 $\pm$ 0.45 & 44.89 $\pm$ 0.36 & 49.50 $\pm$ 0.23 & 53.41 $\pm$ 0.23 & 56.39 $\pm$ 0.05 & 58.10 $\pm$ 0.11 & 59.08 $\pm$ 0.21\\\hline
        SPELA 5x top5 & 52.58 $\pm$ 0.84 & 72.18 $\pm$ 0.22 & 76.02 $\pm$ 0.30 & 79.15 $\pm$ 0.21 & 80.98 $\pm$ 0.22 & 82.32 $\pm$ 0.13 & 82.96 $\pm$ 0.16\\\hline
        BP top1 & 55.34 $\pm$ 0.36 & 58.97 $\pm$ 0.34 & 56.93 $\pm$ 1.30 & 62.16 $\pm$ 0.29 & 62.61 $\pm$ 0.08 & 62.40 $\pm$ 0.16 & 62.09 $\pm$ 0.10\\\hline
        BP top5 & 80.95 $\pm$ 0.26 & 83.66 $\pm$ 0.24 & 82.96 $\pm$ 0.72 & 85.49 $\pm$ 0.07 & 85.71 $\pm$ 0.09 & 85.51 $\pm$ 0.16 & 85.21 $\pm$ 0.09\\\toprule
    \end{tabular}
    \end{adjustbox}
    \caption{Test accuracy for Food 101 dataset.}
    \label{tab:tinyTL accuracies dataset Food}
\end{table}

\begin{table}[H]
    \centering
   
    \begin{adjustbox}{width=1.\textwidth}
    \begin{tabular}{|c| c| c| c| c| c| c| c| c|}
    \hline
      \textbf{Test size:} & \textbf{1} & \textbf{5} & \textbf{10} & \textbf{25} & \textbf{50} & \textbf{75} & \textbf{100}\\\hline
        
        SPELA top1 & 13.07 $\pm$ 2.23 & 23.58 $\pm$ 3.15 & 25.56 $\pm$ 1.81 & 33.48 $\pm$ 0.86 & 40.65 $\pm$ 1.92 & 41.75 $\pm$ 1.58 & 42.95 $\pm$ 1.62\\\hline
        SPELA top5 & 32.28 $\pm$ 2.12 & 45.18 $\pm$ 3.42 & 48.23 $\pm$ 1.62 & 56.06 $\pm$ 1.22 & 61.84 $\pm$ 1.38 & 61.68 $\pm$ 2.01 & 62.51 $\pm$ 1.29\\\hline
        SPELA 5x top1 & 12.52 $\pm$ 1.24 & 16.65 $\pm$ 1.38 & 16.82 $\pm$ 1.20 & 38.65 $\pm$ 0.88 & 55.37 $\pm$ 1.08 & 61.29 $\pm$ 0.80 & 64.82 $\pm$ 0.82\\\hline
        SPELA 5x top5 & 28.89 $\pm$ 2.37 & 35.27 $\pm$ 1.77 & 36.65 $\pm$ 1.52 & 62.67 $\pm$ 0.87 & 76.39 $\pm$ 0.43 & 80.31 $\pm$ 0.85 & 82.40 $\pm$ 1.31\\\hline
        BP top1 & 58.61 $\pm$ 4.33 & 75.62 $\pm$ 0.73 & 77.24 $\pm$ 0.68 & 79.53 $\pm$ 0.32 & 80.12 $\pm$ 0.10 & 80.39 $\pm$ 0.07 & 80.34 $\pm$ 0.27\\\hline
        BP top5 & 82.72 $\pm$ 2.03 & 91.64 $\pm$ 0.33 & 92.21 $\pm$ 0.30 & 93.16 $\pm$ 0.26 & 93.39 $\pm$ 0.28 & 93.52 $\pm$ 0.05 & 93.61 $\pm$ 0.10\\\toprule
    \end{tabular}
    \end{adjustbox}
    \caption{Test accuracy for Flowers 102 dataset.}
    \label{tab:tinyTL accuracies dataset Flowers}
\end{table}

\subsection{Extension of Ablation Studies}
\label{sec:ablation_appendix}

\subsubsection{Relevance of Euclidean distance}
The goal of learning in SPELA is to learn to orient the activation vector $h_{i}$ towards the correct class $c$ symmetric vector $v_{c}$. This is akin to reducing the angle between $h_{i}$ and $v_{c}$, which in turn increases the cosine similarity or angular similarity. Hence, cosine is an appropriate similarity. 

If, for instance, $h_{i} = 100 v_{c}$, then even though the cosine loss would be zero due to perfectly aligned vectors, the Euclidean distance would be high, and hence the corresponding loss would be high. We notice that in such a scenario, SPELA's performance drops significantly, SPELA-Euc in Table \ref{tab:ablation_studies_appendix}.  But if we use normalized Euclidean distance wherein the vectors are $\frac{h_{i}}{||h_{i}||_{2}}$ and  $\frac{v_{c}}{||v_{c}||_{2}}$, then the performance of SPELA closely matches that of cosine, SPELA-Norm-Euc in Table \ref{tab:ablation_studies_appendix}. 

\begin{table}[!htb]
    \centering
    \begin{tabular}{|c| c| c| c| c|}
        \hline
        \textbf{Model} & \textbf{Architecture} & \textbf{MNIST 10} & \textbf{\makecell{KMNIST 10}} & \textbf{\makecell{Fashion MNIST 10}} \\
        \hline 
        SPELA-Cos & $784\rightarrow1024$ &  91.09 $\pm$ 0.12 & 68.74 $\pm$ 0.18 & 84 $\pm$ 0.11\\
        \hline 
        SPELA-Cos & $784\rightarrow1024\rightarrow10$ & 94.41 $\pm$ 0.49 & 75.05 $\pm$ 3.47 & 84.12 $\pm$ 3.30\\
        
        \hline 
        SPELA-Euc & $784\rightarrow1024$ &  75.37 $\pm$ 0.34 & 52.71 $\pm$ 0.28 &  67.83 $\pm$  0.77\\
        \hline 
        SPELA-Euc & $784\rightarrow1024\rightarrow10$ & 48.16  $\pm$ 8.76 & 28.16 $\pm$ 8.14 &  51.82 $\pm$ 7.18 \\

         \hline 
        SPELA-Norm-Euc & $784\rightarrow1024$ &  90.24 $\pm$  0.11 & 67.11 $\pm$ 0.13 &  83.53 $\pm$  0.09\\
        \hline 
        SPELA-Norm-Euc & $784\rightarrow1024\rightarrow10$ & 90.0  $\pm$ 3.64 & 71.57 $\pm$ 2.2 &  84.60 $\pm$ 1.14 \\
        
        \toprule
        
    \end{tabular}
    \caption{Ablation study results of SPELA on MNIST 10, KMNIST 10, and FashionMNIST 10 datasets after $200$ training epochs (and for five runs). SPELA-Cos implies SPELA with cosine distance, SPELA-Euc implies SPELA with Euclidean distance,  SPELA-Norm-Euc implies SPELA with Euclidean distance on normalized activation and symmetric vectors.}
    \label{tab:ablation_studies_appendix}
\end{table}

\subsubsection{Remarks on randomizing the Vector Embeddings}
\label{sec: random_vec_appendix}

\paragraph{Remark 1:}As we operate in dimensions much higher than the number of embedded vectors (number of classes), a non-symmetric distribution should perform equivalent to a symmetric structure. The performance gap between symmetric and non-symmetric structures would be noticeable when the number of dimensions exceeds the number of classes.

\paragraph{Remark 2:}We use the energy of the system as a structured metric:

\[
\lambda(\mathcal{V}) = \sum_{\mathbf{u} \in \mathcal{V}} \sum_{\mathbf{v} \in \mathcal{V}, \mathbf{u}\neq\mathbf{v}} \frac{1}{\| \mathbf{u} - \mathbf{v}\|}
\]

\noindent Of all possible vectors $\mathbf{x}\in\mathbb{R}^d$, if $|\mathcal{V}|$ is fixed, the symmetric structure has the minimum energy. Comparing vectors drawn from the Gaussian distribution $\mathcal{N}(\mathbf{0}, \mathbf{1})$ and the symmetric structure, we get a high energy difference. Despite this, the network learns from the incoming data. A symmetric structure is necessary for lower dimensions (where the number of dimensions is comparable to the number of embedded vectors).

We learn that although having vectors embedded in a symmetric structure is optimal, it is unnecessary. The model will mold the weights according to the relative positions of the vectors and classify the data accordingly, which ascertains the model's flexibility.

\begin{table}[!h]
    \centering
    \begin{tabular}{|c| c| c| c| c| c| c| c|}
        \hline
        \textbf{Dataset} & \textbf{Architecture} & \textbf{0.01} & \textbf{0.1} & \textbf{1} & \textbf{1.5} & \textbf{2.5} & \textbf{3} \\
        \hline 
        MNIST 10 & $784\rightarrow 1024$ & \makecell{ 90.27\\ $\pm$ 1.1 } &  \makecell{  90.97  \\ $\pm$ 0.12 } &  \makecell{ 89.72 \\ $\pm$ 1.01 } &  \makecell{91.26 \\ $\pm$ 0.05} &  \makecell{91.01\\ $\pm$ 0.12} &  \makecell{ 90.97 \\ $\pm$ 0.06 } \\
        \hline 
        MNIST 10 & $784\rightarrow 1024\rightarrow 10$ & \makecell{ 91.86 \\ $\pm$ 3.22 } &  \makecell{  92.92  \\ $\pm$ 1.63} &  \makecell{ 86.93 \\ $\pm$ 11.06 } &  \makecell{ 93.31 \\ $\pm$ 1} &  \makecell{94.25 \\ $\pm$ 1.12} &  \makecell{ 92.31 \\ $\pm$ 3.28} \\
        \hline 
        KMNIST 10 & $784\rightarrow 1024$ &  \makecell{65.74 \\ $\pm$  2.77 } &  \makecell{ 68.54 \\ $\pm$ 0.18 } &  \makecell{ 64.82\\ $\pm$ 3.23 } &  \makecell{ 67.9\\ $\pm$ 0.5 } &  \makecell{ 68.48\\ $\pm$ 0.11 } &  \makecell{ 68.6\\ $\pm$ 0.17 }  \\
        \hline 
        KMNIST 10 & $784\rightarrow 1024\rightarrow 10$ & \makecell{75.06 \\ $\pm$  2.47 } &  \makecell{   74.30  \\ $\pm$ 3.38} &  \makecell{ 73.86 \\ $\pm$3.81  } &  \makecell{66.11\\ $\pm$3.57 } &  \makecell{74.12 \\ $\pm$ 2.41} &  \makecell{ 75.69 \\ $\pm$ 2.03} \\
        \hline 
        Fashion MNIST 10 & $784\rightarrow 1024$ &  \makecell{ 81.57\\ $\pm$ 2.11 } &  \makecell{ 83.97 \\ $\pm$0.06  } & \makecell{81.10 \\ $\pm$ 2.35 } & \makecell{83.34\\ $\pm$ 0.57 } & \makecell{83.99 \\ $\pm$ 0.06 } & \makecell{ 83.96 \\ $\pm$ 0.05 } 
        \\
        \hline 
        Fashion MNIST 10 & $784\rightarrow 1024\rightarrow 10$ & \makecell{ 83.12 \\ $\pm$ 4.51 } &  \makecell{  85.34  \\ $\pm$ 0.21} &  \makecell{81.32\\ $\pm$ 3.28 } &  \makecell{83.65\\ $\pm$ 1.61} &  \makecell{84.52 \\ $\pm$ 1.66} &  \makecell{  81.6\\ $\pm$ 4.99} \\

        \toprule
    \end{tabular}
    \caption{SPELA test accuracies (after $200$ epochs of training and over five runs) for learning rates of 0.01, 0.1, 1, 1.5, 2.5, and 3.}
    \label{tab:spela_lr}
\end{table}

\subsection{Extension of SPELA Convolutional Neural Network}
\label{sec:spelacnn_appendix}

We next varied the channels in SPELA\_B\_CNN from 32 to 128 in SPELA\_C\_CNN and observed that SPELA improves performance on CIFAR 10, CIFAR 100, as well as SVHN 10 (Table \ref{tab:spela_cnn_appendix}).   

\begin{table}[!h]
    \centering
    \begin{center}
    \begin{tabular}{|c|c|c|c|c|c|c|}
        \hline
        \textbf{Model} & \textbf{\makecell{CIFAR 10}} & \textbf{\makecell{CIFAR 100}} & \textbf{\makecell{SVHN 10}}\\
         \hline
        SPELA\_B CNN(2) & 56.59 $\pm$ 0.97 &  26.71 $\pm$ 0.79  & 76.28 $\pm$ 6.45
        \\
         \hline
        SPELA\_C CNN(2) & 64.76 $\pm$ 0.49 &  27.59 $\pm$ 4.7  & 84.93 $\pm$ 0.21
        \\
        \toprule
        
    \end{tabular}
    \end{center}
    \caption{Test accuracies (mean $\pm$ standard deviation) comparison of different SPELA CNN architectures on CIFAR 10, CIFAR 100, and SVHN 10 datasets.  SPELA\_B and SPELA\_C indicate networks with 32 and 128 channels, respectively(Please refer to Section C.4 for experimental details). }
    \label{tab:spela_cnn_appendix}
\end{table}

\section{Experimental Details}

We used an NVIDIA RTX 4500 Ada generation GPU for all our studies. 

\subsection{How does SPELA work?}

\begin{table}[H]
    \centering
    \begin{center}
    \begin{tabular}{|c|c|c|c|}
        \hline
         & \textbf{\makecell{BP\_CH\_A}} & \textbf{\makecell{SPELA\_CH\_A}} & \textbf{\makecell{SPELA\_A}}\\
        \hline
        Layer 1 & 1024 & 1024 & 1024
        \\
        \hline
        Layer 2 & 10 & 10 & 10 
       \\
       \hline
       Learning rate & 0.1 & 0.1 & 0.1
       \\
       \hline
       Decay rate & $\times$ 0.1 & $\times$ 0.1 & $\times$ 0.1
       \\
       \hline
       Decay epoch & 60 & 60 & 60
       \\
       \hline
       Batch size & 50 & 50 & 50
       \\
       \hline
       \# Epochs & 100 & 100 & 100
       \\
       \hline
       Dropout & 10$\%$  & 10$\%$ & 10$\%$
       \\
       \hline
       Weight init & He Uniform & He Uniform & He Uniform
       \\
       \hline
       Bias & False & False  & False 
       \\
       \hline
       Optimizer & SGD & SGD & SGD
       \\
       & (momentum=0.9) & & 
       \\
       \hline
       Activation & ReLU & ReLU & ReLU
       \\
       \hline
       Loss & Cross-Entropy & Cross-Entropy & Positive Cosine
       \\
        \hline
        
    \end{tabular}
    \end{center}
    \caption{Experiment Details for SPELA MLP}
    \label{tab:spela_mlp_exp_details_A}
\end{table}


\begin{table}[H]
    \centering
    \begin{center}
    \begin{tabular}{|c|c|c|c|}
        \hline
         & \textbf{\makecell{BP\_CH\_B}} & \textbf{\makecell{SPELA\_CH\_B}} & \textbf{\makecell{SPELA\_B}}\\
        \hline
        Layer 1 & 1024 & 1024 & 1024
        \\
        \hline
        Layer 2 & 10 & 10 & 10 
       \\
       \hline
       Learning rate & 2.5 & 2.5 & 2.5
       \\
       \hline
       Decay rate & $-$ 0.1 & $-$ 0.1 & $-$ 0.1
       \\
       \hline
       Decay epoch & Every 10 & Every 10 & Every 10
       \\
       \hline
       Batch size & 50 & 50 & 50
       \\
       \hline
       \# Epochs & 200 & 200 & 200
       \\
       \hline
       Dropout & 0 & 0 & 0
       \\
       \hline
       Weight init & He Uniform & He Uniform & He Uniform
       \\
       \hline
       Bias & True & True  & True 
       \\
       \hline
       Optimizer & SGD & SGD & SGD
        \\
       \hline
       Activation &  Leaky ReLU(Slope=0.001) & Leaky ReLU(Slope=0.001) & Leaky ReLU(Slope=0.001)
       \\
       \hline
       Loss & Cross-Entropy & Cross-Entropy & Positive Cosine
       \\
        \hline
        
    \end{tabular}
    \end{center}
    \caption{Experiment Details for SPELA MLP}
    \label{tab:spela_mlp_exp_details_B}
\end{table}

\subsection{Transfer Learning with SPELA}

\begin{table}[H]
    \centering
    \begin{center}
    \begin{tabular}{|c|c|c|c|}
        \hline
         & \textbf{\makecell{BP}} & \textbf{\makecell{SPELA}} & \textbf{\makecell{SPELA 5x}}\\
        \hline
        Layer 1 & \#Classes & \#Classes & 5 $\times$ \#Classes
        \\
       \hline
       Learning rate & 0.1 & 0.1 & 0.1
       \\
       \hline
       Decay rate & 0 & 0  & 0
       
       \\
       \hline
       Batch size & 128 & 128 & 128
       \\
       \hline
       \# Epochs & 200 & 200 & 200
       \\
       \hline
       Dropout & 0 & 0 & 0
       \\
       \hline
       Weight init & He Uniform & He Uniform  & He Uniform 
       \\
       \hline
       Bias & True & True & True 
       \\
       \hline
       Optimizer & SGD & SGD & SGD
        \\
       \hline
       Loss & Cross-Entropy & Cross-Entropy & Cross-Entropy
       \\
        \hline
        
    \end{tabular}
    \end{center}
    \caption{Experiment Details of SPELA on Transfer Learning}
    \label{tab:spela_mlp_transfer_learning_expdetails}
\end{table}

\subsection{Ablation Studies}

\begin{table}[H]
    \centering
    \begin{center}
    \begin{tabular}{|c|c|c|}
        \hline
         & \textbf{\makecell{BP}} & \textbf{\makecell{SPELA}}\\
        \hline
        Layer 1 & 1024 & 1024
        \\
        \hline
        Layer 2 & 10 & 10 
       \\
       \hline
       Learning rate & 2.5 & 2.5
       \\
       \hline
       Decay rate & $-$ 0.1 & $-$ 0.1
       \\
       \hline
       Decay epoch & Every 10 & Every 10
       \\
       \hline
       Batch size & 50 & 50
       \\
       \hline
       \# Epochs & 200 & 200
       \\
       \hline
       Dropout & 0 & 0
       \\
       \hline
       Weight init & He Uniform & He Uniform
       \\
       \hline
       Bias & True  & True 
       \\
       \hline
       Optimizer & SGD & SGD
        \\
       \hline
       Activation & Leaky ReLU(Slope=0.001) & Leaky ReLU(Slope=0.001)
       \\
       \hline
       Loss & Cross-Entropy & Positive Cosine
       \\
        \hline
        
    \end{tabular}
    \end{center}
    \caption{Experiment Details of SPELA MLP for Ablation Studies}
    \label{tab:spela_mlp_ablation_exp_details}
\end{table}

\subsection{SPELA Convolutional Neural Network}

\begin{table}[H]
    \centering
    \begin{center}
    \begin{tabular}{|c|c|c|}
        \hline
         & \textbf{\makecell{SPELA\_CH\_B}} & \textbf{\makecell{SPELA\_B}}  \\
        \hline
         Input size & 32×32×3 & 32×32×3
        \\
        \hline
         Conv & 32,5,1(2) & 32,5,1(2)\\
         \hline
         MLP & 10/100 & 10/100
       \\
       \hline 
       Learning rate & 0.1, 0.1 & 0.1, 0.1
       \\
       \hline
       Decay rate & 0 & 0
       
       \\
       \hline
       Batch size & 64 & 64 
       \\
       \hline
       \# Epochs & 15+10 & 15+10 
       \\
       \hline
       Dropout & 0 & 0 
       \\
       \hline
       Weight init & Kaiming Uniform & Kaiming Uniform 
       \\
       \hline
       Bias & True  & True 
       \\
       \hline
       Optimizer & Adam & Adam 
        \\
       \hline
       Activation &  PReLU,Leaky ReLU(Slope=0.001)  &  PReLU,Leaky ReLU(Slope=0.001) 
       \\
       \hline
       Loss & Cross-Entropy & Positive Cosine
       \\
        \hline
        
    \end{tabular}
    \end{center}
    \caption{Experimental details of SPELA convolutional neural network.}
    \label{tab:spela_cnn}
\end{table}

\end{document}

%% file: math_commands.tex
\usepackage{amsmath,amsfonts,bm}

\def\eqref#1{equation~\ref{#1}}

\def\1{\bm{1}}

\DeclareMathAlphabet{\mathsfit}{\encodingdefault}{\sfdefault}{m}{sl}
\SetMathAlphabet{\mathsfit}{bold}{\encodingdefault}{\sfdefault}{bx}{n}

\DeclareMathOperator*{\argmax}{arg\,max}